\documentclass[aps,pra,superscriptaddress,twocolumn,longbibliography]{revtex4-1}
\usepackage{mathrsfs}
\usepackage{amsfonts}
\usepackage{amssymb}
\usepackage{amsmath}
\usepackage{graphicx}
\usepackage{sidecap}
\usepackage{color}
\usepackage[colorlinks=true, urlcolor=blue, citecolor=blue,linkcolor=blue,citebordercolor={1 0 0},linkbordercolor={0 0 1}]{hyperref}
\usepackage{subeqnarray}
\usepackage{euscript} 
\usepackage{hyperref}

\usepackage{array}
\usepackage{multirow}

\usepackage{shortcuts}
\usepackage{bm}
\usepackage{color}
\usepackage{amsthm,parskip,setspace,tabularx, wrapfig,float,enumerate} 
\usepackage{tikz, setspace, subfig, grffile}
\usepackage{subfig}
\usepackage{paralist}
\usepackage{float}
\usepackage[ruled,vlined]{algorithm2e}
\usepackage[]{algpseudocode}

\newcommand{\ket}[1]{\vert #1 \rangle}
\newcommand{\bra}[1]{\langle #1 \vert}

\newcommand{\braket}[2]{\langle #1 \vert #2 \rangle}

\renewcommand{\eqref}[1]{Eq.~(\ref{#1})} 
\usepackage[columnwise]{lineno} 

\newcommand{\utchem}{Department  of  Chemistry,  University  of  Toronto,  Toronto,  Ontario  M5G 1Z8,  Canada}
\newcommand{\utcomp}{Department  of  Computer Science,  University  of  Toronto,  Toronto,  Ontario  M5S 2E4,  Canada}
\newcommand{\vectorinst}{Vector  Institute  for  Artificial  Intelligence,  Toronto,  Ontario  M5S  1M1,  Canada}
\newcommand{\cifar}{Canadian  Institute  for  Advanced  Research,  Toronto,  Ontario  M5G  1Z8,  Canada}
\newcommand{\xumei}{Hefei National Laboratory for physical Science at Microscale and Department of Modern Physics, University of Science and Technology of Chine, Hefei, Anhui 230026, China,}

\newcommand{\mpl}{Max Planck Institute for the Science of Light (MPL), 91058 Erlangen, Germany}

\newcommand{\red}[1]{\textcolor{black}{#1}}

\begin{document}
\title{Learning Interpretable Representations of Entanglement\\ in Quantum Optics Experiments using Deep Generative Models }

\author{Daniel Flam-Shepherd}
\email{danielfs@cs.utoronto.edu}
\affiliation{\utcomp}
\affiliation{\vectorinst}

\author{Tony Wu}
\affiliation{\utcomp}

\author{Xuemei Gu}
\affiliation{\xumei}

\author{Alba Cervera-Lierta}
\affiliation{\utcomp}
\affiliation{\utchem}

\author{Mario Krenn}
\email{mario.krenn@mpl.mpg.de}
\affiliation{\utcomp}
\affiliation{\utchem}
\affiliation{\vectorinst}
\affiliation{\mpl}

\author{Al\'an Aspuru-Guzik}
\email{alan@utoronto.edu}
\affiliation{\utcomp}
\affiliation{\utchem}
\affiliation{\vectorinst}
\affiliation{\cifar}


\begin{abstract}
Quantum physics experiments produce interesting phenomena such as interference or entanglement, which are core properties of numerous future quantum technologies. 
The complex relationship between the setup structure of a quantum experiment and its entanglement properties is essential to fundamental research in quantum optics but is difficult to intuitively understand. 
We present a deep generative model of quantum optics experiments where a variational autoencoder is trained on a dataset of quantum optics experimental setups. In a series of computational experiments, we investigate the learned representation of our Quantum Optics Variational Auto Encoder (QOVAE) and its internal understanding of the quantum optics world. 
We demonstrate that the QOVAE learns an interpretable representation of quantum optics experiments and the relationship between experiment structure and entanglement. We show the QOVAE is able to generate novel experiments for highly entangled quantum states with specific distributions that match its training data. The QOVAE can learn to generate specific entangled states and efficiently search the space of experiments that produce highly entangled quantum states. Importantly, we are able to interpret how the QOVAE structures its latent space, finding curious patterns that we can explain in terms of quantum physics. The results demonstrate how we can use and understand the internal representations of deep generative models in a complex scientific domain. The QOVAE and the insights from our investigations can be immediately applied to other physical systems.
\end{abstract}

\maketitle


\section*{Introduction}

Quantum mechanics contains a wide range of phenomena that seem counter intuitive from a classical physics perspective.  
Experimental quantum physics is integral to the investigation of the fundamental questions associated with these phenomena and the quantum mechanical nature of the universe. 
Quantum entanglement \cite{schrodinger1935discussion,einstein1935can,Bell_1964} is one of those phenomena that is most difficult to reconcile with our picture of reality and also provides the basis for all quantum technologies and applications. Thus, in particular, quantum optics experiments are not only used to test the foundations of quantum physics \cite{giustina2015significant,shalm2015strong,bong2020strong},  they are also at the heart of numerous quantum technologies in many areas including communication \cite{yin2017satellite} and computation \cite{peruzzo2014variational,paesani2019generation,zhong2020quantum}. The quantum optics experiments we consider here consist of individual optical elements or devices, such as lasers, beam splitters, or non-linear crystals.
Complex quantum phenomena such as multi-photon interference effects \cite{wang1991induced,herzog1994frustrated,menssen2017distinguishability,feng2021observation}, are challenging to understand intuitively. For that reason, in general the connection between experimental structures and its entanglement properties -- the so-called structure-property relation -- is complicated to grasp for humans, which leads to undiscovered potential of these technologies.

In order for the continuing advancement of fundamental research and quantum technologies, 
it is advantageous that researchers develop computational methods that help in the designing of new quantum hardware while providing conceptual understanding of the results \cite{krenn2020computer}. Examples include the Melvin algorithm that learns to expand its own toolbox with useful elements \cite{Krenn_2016}, or a graph-based topological optimizer that allows to extract new human-interpretable concepts \cite{krenn2020conceptual}. Other works show how to optimize setups with genetic algorithms \cite{knott2016search,nichols2019designing,o2019hybrid}, reinforcement learning \cite{Melnikov_2018} or parametrized optimization \cite{arrazola2019machine}. 
These efforts do not directly generate quantum optics experiments through the use of a learned representation trained on examples of experiments. Such an approach would provide us with the ability to generate with prior knowledge of specific entangled experiments and allow us to directly explore the relationship between experiment structure and entanglement in model's learned representation. Therefore, in this work, we focus on using deep unsupervised learning \cite{salakhutdinov2015learning} and build a generative model of quantum optics experiments.  

Deep generative models have had a major impact in the past few years where they have been applied successfully to a variety of data, including images \cite{razavi2019generating}, text \cite{bowman2015generating,semeniuta2017hybrid} and audio \cite{roberts2018hierarchical}. Generative models allow one to generate new examples similar to the training data. In particular, many advances have been made using deep generative models in the chemical sciences \cite{sanchez2018inverse}, for example, Variational autoencoders (VAEs) \cite{kingma2013auto} have been widely used for molecular design \cite{gomez2018automatic,samanta2020nevae, jin2018junction, flam2021mpgvae, jin2020hierarchical, yao2021inverse}. They enable us to generate specific distributions of molecules with certain molecular properties in order to efficiently search through chemical space \cite{gomez2018automatic,liu2018constrained}. This offers advantages over other approaches that generate molecules without prior knowledge of the targeted distribution. In particular, VAEs allow for the efficient optimization of discrete molecular structures by learning continuous latent representations.

Learning interpretable representations \cite{bengio2013representation,higgins2016beta} of generative factors of structured data in science is an important precursor for the development of artificial intelligence that is able to learn concepts in order to make scientific discoveries \cite{Iten_2020,higgins2016beta}. For example, in a supervised setting-- SciNet \cite{Iten_2020} has been used to gain conceptual insights using times series observations from simple physical systems, motivated by this, we investigate learning interpretable representations in a unsupervised way with structured data in a complex scientific domain.

Specifically we study if deep generative models can learn representations of entangled experiments in an intrepretable way in order to efficiently explore the space of quantum optics experiments. 
We demonstrate that our model, the QOVAE-- the first deep generative model of quantum optics experiments, can learn a representation that encodes the relationship between entanglement and experiment structure that enables it to generate diverse and novel setups from distributions of high dimensionally entangled quantum optics experiments. 
We find that the QOVAE is able to generate very specific spaces of entangled states by training on tightly constrained ranges of entanglement in experimental setups, in particular, the QOVAE can generate the most challenging training experiments-- with highest entangled and least devices-- faster than it takes random sampling to produce such experiments.
We further demonstrate a method using bayesian optimization in the QOVAE's latent space to search and target individual states using a novel objective that can be customized with specific constraints according to the scientist's interest.
Most importantly we show that the QOVAE can learn an interpretable representation of experiments-- by doing so we discover how the model learns-- opening its black-box to see the QOVAE encodes experiments in its latent space according to each experiment's length and ordering of devices.
Our central contribution is not solely that the QOVAE can be used for experiment design-- which many very powerful domain specific methods already exist \cite{krenn2020conceptual,krenn2020computer}
but rather that the QOVAE can autonomously learn highly complex systems in a human-intrepretable way that could be understood and potentially practically exploited by scientific experts for the investigation of highly entangled quantum systems.

\begin{figure*}[t] 
\includegraphics[width=0.9\textwidth]{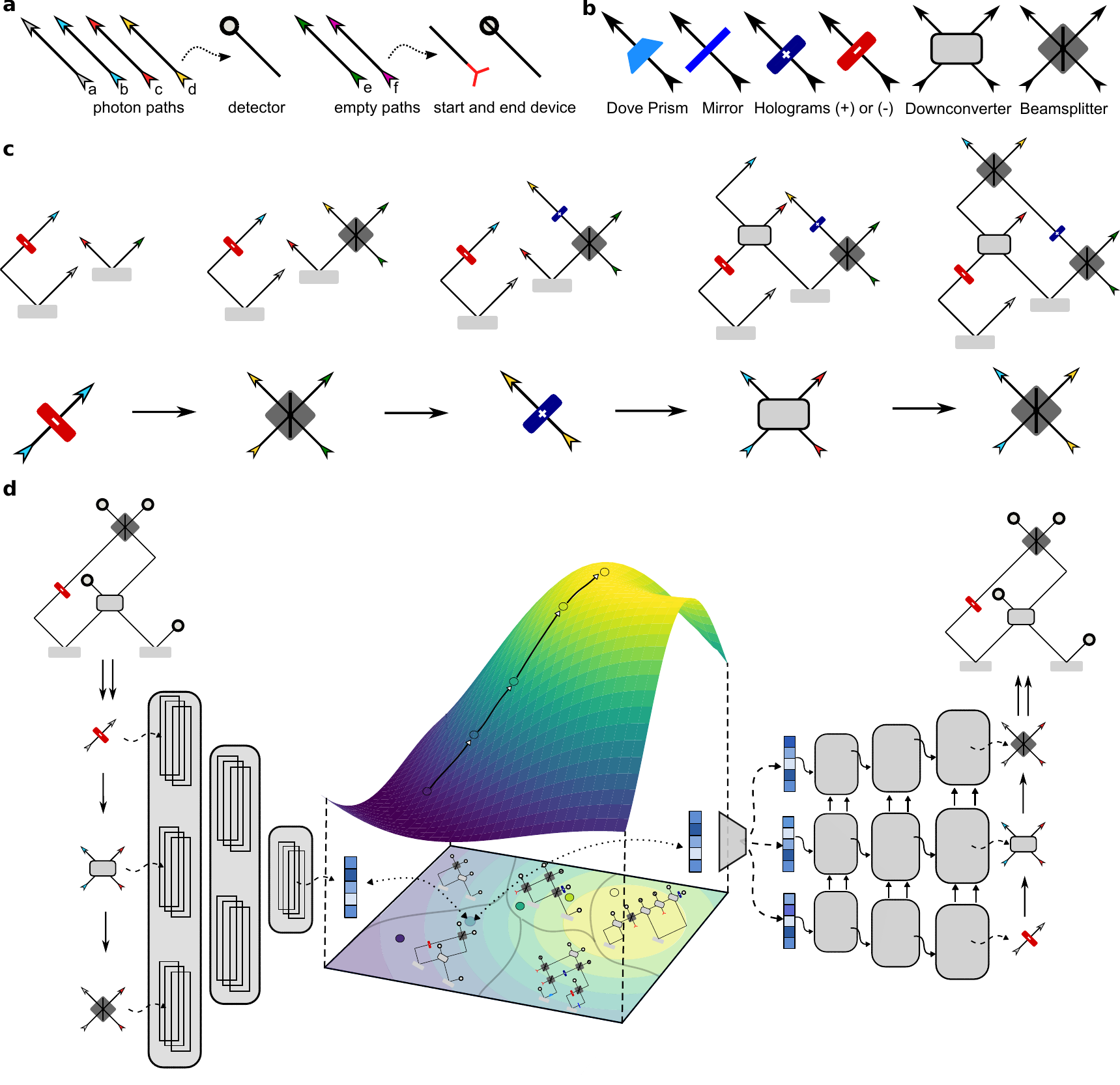} 
\caption{\raggedright \textbf{Data representation and Model.} \textbf{a} Every experiment involves four photons paths beginning with an SPDC crystal to then end with a detector represented by a grey circle with black outline. The paths are designated by arrows and color coded with blue, grey, red and yellow for photons a,b,c,d respectively. In addition, there can be two empty paths e,f (green and purple arrow heads). Empty photon paths start without a crystal and are shown with a red, three pointed star and end with no detector (the detector symbol with a slash).
\textbf{b} A visualization of the toolbox of possible devices in any experiment.
\textbf{c} An example of a quantum optics experiment. Visually shown as a sequence of graphs then below as each device's visualization from the toolbox.   
\textbf{d} A visual depiction of the QOVAE. First, an experiment encoded as a sequence into a stochastic latent representation using a convolutional neural network and is reconstructed using another deep recurrent neural network. The entanglement measure is shown above as a function of the latent space. On the entanglement function arrows are shown moving from a region of low entanglement in violet towards region of high entanglement in yellow.
}
\label{fig:intro}
\end{figure*}

\textbf{Quantum optics experiments}\quad To represent each of the experimental setups, we use a discrete sequence of optical devices as shown in Fig. \ref{fig:intro}\textbf{c}. Every optical device is identified by its location in the graph, specified by the photons propagating through the device and its order in the sequence. Therefore each sequence uniquely determines the final quantum state and entanglement properties of the system.
The quantum system in each experiment is a four photon system with its initial state created by a double spontaneous parametric down-conversion process (SPDC) that experimentally generates two photon pairs.
These SPDC processes can produce multipartite entanglement \cite{Bouwmeester_1999, yao2012observation}, high-dimensional entanglement encoded in the intrinsic orbital angular momentum (OAM) of photons \cite{Allen_1992, Romero_2012, Krenn_2014}, and combinations thereof \cite{erhard2018experimental,luo2019quantum}. OAM is the component of angular momentum dependent on the field spatial distribution.

\textbf{The device toolbox.}\quad The experiments are generated using a set of basic elements consisting of
beam splitters ($\text{BS}_{pp'}$), mirrors ($\text{R}_{p}$), dove prisms ($\text{D}_{p}$), single mode OAM down-converters ($\text{DC}_{pp'}$) and holograms ($\text{H}_{p}^n, n \in \mathbb{Z} $,) \cite{Leach_2002}. 
For each device its operator is sub-scripted by the path(s) it acts on : either a single path $p$ or two paths $p, p'$. 
The holograms and the dove prisms have discrete parameters corresponding to the OAM and phase added to the beam, respectively.
We use a toolbox of 6 kinds of devices operating on 4 possible photon paths with up to 2 empty paths (Fig. \ref{fig:intro}\textbf{a,b}). Empty paths are important for increasing the diversity of states possible. The methods section has more details on the device toolbox and how different devices change the quantum state.

\textbf{Entanglement measure.}\quad The system we study is a high-dimensional four-photon quantum state. To quantify its entanglement, we derive the entanglement entropy from the discrete Schmidt Rank Vector (SRV) \cite{huber2013structure}. The SRV is a vector composed of the Schmidt ranks of all bipartitions which, in the case of four particles, has a size of seven. For an overall measure of entanglement we use $S$: the sum of all bipartition entanglement entropies, where an experiment with $S>0$ is entangled and with $S=0$ is unentangled. The state and entanglement is calculated numerically using the symbolic algebra python package \texttt{sympy} \cite{meurer2017sympy}. In general, computing the state of highly entangled experiments can be expensive-- therefore it is helpful to find more direct methods to find experiments of certain states, towards this goal-- Bayesian optimization in the QOVAE's latent space can be used to search for specific states.

\section*{QOVAE Results} 

\textbf{Model description.}\quad For our QOVAE model, we use a variational autoencoder to learn distributions of quantum experiments as sequences. The QOVAE model consists of two neural networks: an encoder which maps a quantum optics experiment $ \mathbf x$ to a continuous latent representation $\mathbf z$ and decoder that reconstructs the experiments $ \mathbf x$ from the latent representation $ \mathbf z$. Both the encoder and decoder are parameterized by deep neural networks. Fig. \ref{fig:intro}\textbf{d} displays the main model. The encoder of the QOVAE learns a representation by using layers of 1D convolutions that are used to generate the mean and log standard deviation of the latent space. The decoder uses the latent representation of the experiment to generate the experimental sequence using a recurrent neural network.

\textbf{Setup encoding.}\quad For the training data, we represent an experiment sequentially as a series of one-hot column vectors $\mathbf x_t$ in a matrix where $ \mathbf x  = [\mathbf x_1 , \dots , \mathbf x_t , \dots , \mathbf x_T ] \in \mathbb{R} ^{T\times D} $. 
Here, $T$ is maximum experiment length (number of devices) and $D$ is the number of devices in the toolbox.
In any experiment, every possible device on any path or path combination is represented as a one-hot vector $\mathbf x_t \in \{0,1\}^D$.
For example, the experiment in Fig. \ref{fig:intro}\textbf{c} given as a sequence of operators: $\text{H}_{b}^{-1} \to \text{BS}_{de} \to \text{H}_{c}^{1} \to \text{DC}_{bc} \to \text{BS}_{bd} $ would have five one hot vectors for each device (and would be padded with $T-5$ zero vectors $\mathbf 0$).

\textbf{Training data preparation.}\quad We use the Melvin computer algorithm \cite{Krenn_2016} with a fixed device set (essentially random search), to generate a training dataset of quantum optics experiments. 
This involves repeatedly random sampling experimental setups and evaluating them. To randomly sample an experiment with $\ell$ devices, we first sample $\ell \sim \text{uniform}\{3,\dots,T\}$ the experiment length from a discrete uniform distribution over possible lengths 3 to $T$ and then sample what each device $d$ is in the sequence  $ d  \sim \text{uniform}\{\text{BS}_{ab},\text{DC}_{ab} , \text{H}_{a}^{1}, \dots   \}$ from another discrete uniform over the entire device toolbox. Next, we calculate the total entanglement $S$ of the experiment: those with ($S>0$) produce entangled states and ones with ($S=0$) are unentangled. In total, we generate ~200K (thousand) entangled and unentangled setups.

\textbf{The space of entangled experiments.}\quad There is a important distinction between Haar random states \cite{hamma2012quantum} and our random quantum optics experiments created by randomly assembling optical devices from a toolbox-- similar to Melvin \cite{Krenn_2016, Melnikov_2018, adler2019quantum}. These random experiments are not guaranteed to create entangled states, based on this device toolbox-- 
experiments with too few 2-photon devices and specific device orderings can produce a state with a single basis ket or no state at all (both are $S=0$). 

Let $n_{tp}$ be the number of beamsplitters or down-converters in a experiment (two photon devices). 
Higher Entanglement (larger $S$) is more likely with larger experiments and larger $n_{tp}$. 
Indeed, a necessary but not sufficient condition for entanglement is for the experiment to satisfy $n_{tp}>1$.
One can increase the probability of sampling two photon devices to increase $S$ but this is challenging to balance with sampling other devices in order to ensure sampling diverse states.
Most importantly, the experiment's device order exactly determines its entanglement and different orderings will likely produce a different entanglement $S$. 

We can estimate the size of the space of entangled vs unentangled experiments through random sampling. 
We do this for experiments with $\ell=6$ and $3 \leq \ell\leq 12$, sampling exactly the same as when building the training data. 
The results show that entangled experiments make up $33.1\pm 4.0$ \% and $40.6\pm 4.8$\% of the two spaces respectively.  
We report the average of 5 runs of sampling 10K setups $\pm$ the standard error.

\begin{figure*}[t]
\includegraphics[width=\textwidth]{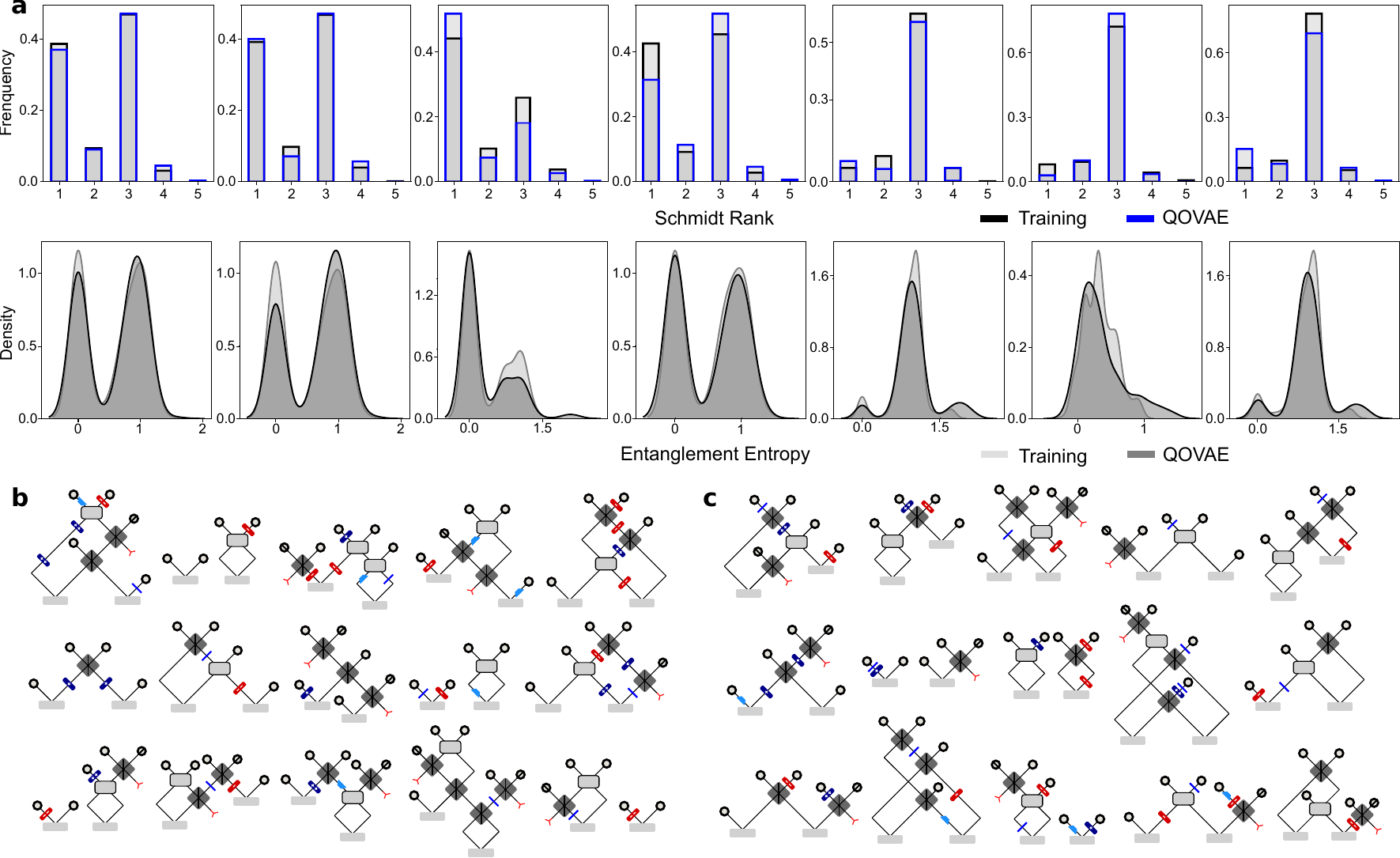}
\caption{\raggedright \textbf{Comparison with Training Data.} \textbf{a} Distribution plots for the entanglement entropy and Schmidt rank of all seven system bi-partitions calculated using experiments generated by the QOVAE-High and training experiments \textbf{b} Training experiments and \textbf{c} experiments generated by the QOVAE-High}
\label{fig:ed}
\end{figure*}
\begin{figure*}[t]
\includegraphics[width=\textwidth]{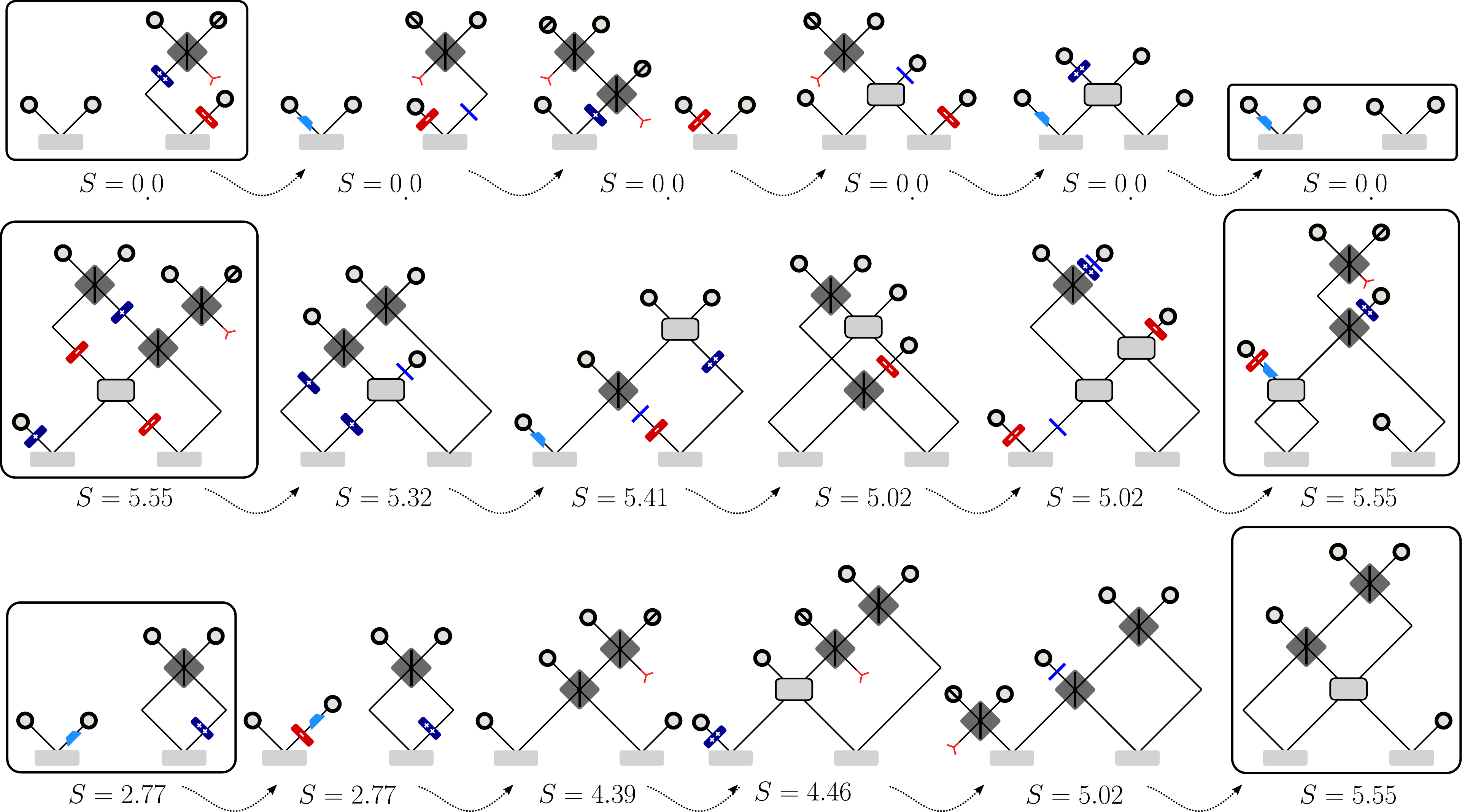}
\caption{\raggedright \textbf{Interpolations.} Three examples of latent space interpolations from the QOVAE. 
Each interpolation is between experiments with different entanglement measure $S$. 
The chosen initial and final experiment along the interpolation path are enclosed in a rectangle and four experiments along the path in between the two are displayed. Below each experiment is the entanglement measure $S$. }
\label{fig:interps}
\end{figure*}

\textbf{Investigations.}\quad For our investigations, we train models on multiple different subsets of the compiled dataset: 
using either the QOVAE-High with a 6 dimensional latent space or the QOVAE-Low with a 2 dimensional latent space.
For different investigations and datasets we train on, we restrict the length $\ell$ of any setup and 
the total number of beamsplitters or down-converters : $n_{tp}$. 
We conduct a number of investigations to assess the QOVAE and its learned representation with different training data, in each, we generate 10K quantum optics experiments from trained models to investigate the entanglement properties. 
Further details about the model can be found in the methods section. 
From the investigations, we extract a series of results that we discuss in the following paragraphs with important conclusions in bold.

\textbf{1) The QOVAE can generate novel experiments from the space of entangled or unentangled quantum optics experiments.} 

In this investigation, we study how capable the QOVAE is in generating from specifics spaces of quantum optics experimental setups that produce entangled and unentangled states. 
We train the QOVAE-High, on experiments with larger lengths $3 \leq \ell \leq 12$ and then small ones with $\ell=6$ (the spaces explored before). 

For each length restriction we train three different QOVAE-High models on 1) unentangled setups, 2) entangled setups and 3) a mixture of the two. 
To ensure that the QOVAE isn't just counting devices to distinguish entangled and unentangled experiments we ensure that both have a similar distribution of two photon devices (responsible for entanglement). This is achieved by ensuring that $n_{tp}>1$ (that each setup has 2 or more two photon devices). This is shown in the supplementary for $\ell\leq 12$. 

From the results we find that the QOVAE produces a similar \% of experiments with $S>0$ that exists in each dataset it is trained on. 
For $\ell=6$, when the QOVAE is trained on unentangled and entangled experiments-- it generates $4.4\pm 2.3$ and  $89.4\pm 3.3$ \% $S>0$ respectively, as well as $50.7\pm 1.2$  when trained on a $48.1 $ mixture of \% $S>0$ and $S=0$.
For $\ell\leq 12$, when the QOVAE is trained on unentangled and entangled experiments-- it generates $7.3 \pm 4.3$ and  $91.5\pm 2.6$ \% $S>0$ respectively, as well as  $52.3\pm 5.5$  when trained on a $50.9\pm 0.0$ \% mixture of  $S>0$ and $S=0$.
We report the average \% of 5 sets of generated experiments $\pm$ the standard error.
For the entangled data, the \% $S>0$ achieved are comparable to the validity scores of generative models of molecules \cite{li2018learning}. 

For these models (and subsequent investigations), we also find that the QOVAE generates 99\% unique experiments-- essentially producing no duplicate experiments. 
As well, it generates all novel experiments that do not appear in the training data. 

Based on these results, we observe that the QOVAE can learn to generate from the space of quantum optics experiments with different entanglement properties.

\textbf{2) The QOVAE can learn distributions of entangled states that it is trained on.} 

Fig. \ref{fig:ed}\textbf{b} displays 15 random samples of experiments from QOVAE-High and its training setups with lengths $\ell\leq 12$ and $0<S\leq 7.5$. We can see the model has learned to generate experiments that have similar structure to the training experiments as both sets of samples have similar numbers of one and two photon devices as well as empty path devices. 

We test if the distribution of entanglement of every bipartition is similar between experiments from the training data and the model. We generate experiments from the model and take training data and calculate all their entanglement entropies and Schmidt ranks in all seven bipartitions. We compare the distributions visually with a distribution plot for each of the seven bipartitions-- we use kernel density estimators \cite{scott2015multivariate} to estimate the entanglement entropy densities and histograms for the Schmidt ranks (Fig. \ref{fig:ed}\textbf{a}). 

We can see from the distribution plots in Fig. \ref{fig:ed}\textbf{a} that in the first four bipartitions (columns) the QOVAE-High successfully learns that the training entanglement has two modes (or peaks). For the last three bipartitions plots (last three columns), the QOVAE-High is able to learn that there is a single mode of entanglement in the training distribution. We notice that the QOVAE learns a heavier right tail than the training distribution.

From the results in Fig. \ref{fig:ed} we conclude that the QOVAE has learned to match the training distribution of entanglement for every bipartition of the system and thus can learn distributions of entangled states.

\textbf{3) The QOVAE learns a quasi-continuous embedding in terms of entanglement.}

We demonstrate the smoothness of the latent space in terms of entanglement $S$ by testing if experiments that are close in the latent space have similar entanglement. 
First we perform spherical interpolations \cite{shoemake1985animating} from one latent representation $\mathbf z_1$ of an experiment to another $\mathbf z_2$, decoding four experiments, at equally spaced steps on the interpolated path from $\mathbf z_1 \to \mathbf z_2$. 

We show three latent space interpolations in Figure \ref{fig:interps}, from different entanglement measures $S$. 
The first interpolation shown in FIG \ref{fig:interps} interpolates between experiments that do not produce entangled states-- we encode two experiments with $S=0$ and generate experiments along the latent path between them. 
In this interpolation, the experiments decoded along the interpolation path remain in the unentangled ($S=0$) space like the initial and final experiments. 

The the second is between experiments that both have $S=5.55$ and the last between one experiment with $S=2.77$ and another with $S=5.55$.  In the second interpolation, 
decoded experiment entanglement is within 0.5 of the initial and final experiments' ($S=5.5$). 
In the last interpolation, the entanglement $S$ of setups decoded along the path increases linearly from $S=2.77$ towards the final experiment's entanglement of $S = 5.55 $.

We also evaluate if nearby experiments in the latent space have similar entanglement properties by comparing the Euclidean distance between latent points and their absolute entanglement difference. We find that as the latent distance increases-- the absolute entanglement difference increases and when the latent distance goes to 0-- the difference in entanglement does as well. A plot of this relationship obtained by random sampling is shown in the supplementary.
Thus, nearby experiments in the latent space are more likely to have similar entanglement properties than experiments further away.

Hence, the QOVAE learns a representation that encodes a measure of similarity between experiment and entanglement $S$.

\begin{table*}[t]
\caption{\raggedright \textbf{Learning different levels of entanglement.} The mode (the value most likely to be sampled), mean and standard deviation (SD) from samples of generated experiments from the QOVAE trained on experiments from four different levels of entanglement. Compared with the real mode, mean and SD of the training data. We use the average over five generations of experiments along with the standard error.}
\begin{tabular}{c|cc|cc|cc|cc}
        \multicolumn{1}{c}{} & \multicolumn{2}{c}{$2.0<S<3.0$ }    &\multicolumn{2}{c}{$3.0<S<4.0$ }  &\multicolumn{2}{c}{$4.0<S<5.0$ }& \multicolumn{2}{c}{$5.0<S<6.0$} \\ 
        Metric   &   QOVAE            & TRAIN             & QOVAE             & TRAIN             & QOVAE             & TRAIN             &  QOVAE            & TRAIN \\ \hline \hline
        Mode $S$ &  $ 2.77 \pm 0.00 $ & $ 2.77 \pm 0.0 $  & $ 3.47 \pm 0.00 $ & $ 3.47 \pm 0.0 $ & $ 4.39 \pm 0.00 $ & $ 4.39 \pm 0.0 $ & $ 5.55 \pm 0.00 $ & $ 5.55 \pm 0.00 $\\
        Mean $S$ &  $ 3.05 \pm 0.45 $ & $ 2.75 \pm 0.0 $  & $ 3.88 \pm 0.42 $ & $ 3.49 \pm 0.0 $ & $ 4.59 \pm 0.46 $ & $ 4.37 \pm 0.0 $ & $ 5.71 \pm 0.55 $ & $ 5.33 \pm 0.00 $\\
        SD $S$   &  $ 0.80 \pm 0.33 $ & $ 0.12 \pm 0.0 $  & $ 1.03 \pm 0.52 $ & $ 0.13 \pm 0.0 $ & $ 0.89 \pm 0.26 $ & $ 0.21 \pm 0.0 $ & $ 1.10 \pm 0.31 $ & $ 0.24 \pm 0.00 $\\
\hline
\end{tabular}
\label{tab:class}
\end{table*}

\textbf{4) The QOVAE can be used to efficiently search for new highly entangled experiments and states.}

We find that the QOVAE can efficiently search for new experiments with higher entanglement than random search 
and that the QOVAE is able to target and generate from specific levels of entanglement 
as well it can generate the most entangled training experiments with fewest devices, far faster than the random methods that produced them.
Additionally, bayesian optimization can be used in the QOVAE's latent space to exactly target specific states.

We report that the QOVAE can learn precise distributions of different levels of entanglement, defined by small intervals of entanglement with unity length. We train the QOVAE on four datasets with entanglement levels : 1) $2.0<S<3.0$, 2) $3.0<S<4.0$, 3) $4.0<S<5.0$ and 4) $5.0<S<6.0$ (using 14K, 11K, 18K and 10K setups respectively). The results in TABLE \ref{tab:class} show for each level of $S$, the QOVAE perfectly matches the training mode of $S$ (the $S$ value most likely to be sampled) and learns the training mean $S$ within 1 standard error but over-estimates the $SD$ of $S$ -- confirming that the QOVAE seems to learn heavier right-tails for $S$ (Figure \ref{fig:ed}\textbf{a}). The model can perfectly capture the mode with no error. This is expected for a well trained QOVAE, while $S$ is continuous the training setups don't all have different values -- in fact one prominent value of $S$ dominates. 
For example, in the level $4.0 < S < 5.0$, there are 677 different $S$ values in the 18K training setups but 40\% of the 18K have $S=4.39$.
Each level of entanglement, basically defines a group of  similar states-- so the QOVAE can learn to target and generate states in these groups.

Next we train the QOVAE-high on the most difficult training experiments to find: some 5K out of 400K total that have the highest entanglement ($S>8.0$) and simplest structure (10 or less devices). 
The results show the QOVAE can learn to directly generate from this space as almost all-- $ 85.4 \pm 3.64 $ \% of generated experiments have $S>8.0$ (and 10 or less devices). 
It also learns to produce the same mode ($S=8.79$) as the training data and within 1 standard error of the mean entanglement ($ 8.55 \pm 0.35 $ for  $ 8.84$). 
Most importantly, using the QOVAE we can generate almost twice the number of these experiments with highest training entanglement and least number of devices -- 
at a small fraction of the time (hours instead of days) it took to produce these experiments when building the training dataset. 
This demonstrates an important advantage of the QOVAE's prior knowledge and is especially useful given how expensive entanglement is to calculate.

We can also target specific states by performing Bayesian optimization in the latent space, taking inspiration from \cite{gomez2018automatic, kusner2017grammar}, we use the target objective defined as $y(s) = |\langle \psi_* |\psi_s\rangle |^{2} - \lambda \text{length}(s)/4d $. This is the fidelity between the target state $|\psi_*\rangle$ and the state $|\psi_s\rangle$ of experimental setup $s$ penalized by experiment length ($d$ is the max experiment length) and $\lambda$ is parameter that can be tuned in order to strengthen or weaken the device penalty. To perform Bayesian optimization (BO), we first train the QOVAE so that each training experiment has a latent vector defined by the mean of the encoder. 
After, we train a sparse Gaussian process to predict $y(s)$ given its latent representation. 
Then we perform a set number of iterations of batched BO using the expected improvement heuristic-- compiling the top scoring states.

As a toy example, we target the 2-dimensional 4 photon GHZ state 
$$|\psi_* \rangle = \frac{|0000\rangle +|1111\rangle}{\sqrt{2}}$$
which is the state where each bipartition has Schmidt rank of 2 and entropy of 0.693-- for a total entanglement measure of $S=4.852$. We set the device penalty very small using $\lambda=0.1$ and run BO for five iterations. We find a single quantum optics experiment that has the exact targeted state ($|\langle \psi_* |\psi_s\rangle |^{2}=1$). This experiment has 11 devices-- 3 reflection devices, 5 beamsplitters, 3 holograms and 1 dove prism that are in the following sequence :
$\text{R}_{a}\to \text{H}_{d}^{-1}\to \text{BS}_{bc}\to  \text{D}_{d}\to \text{R}_{c}\to \text{H}_{b}^{-1}\to \text{R}_{d} \to \text{BS}_{ab} \to \text{BS}_{cd} \to \text{BS}_{ac}\to \text{BS}_{ac} $. 

\begin{figure*}[t]
\includegraphics[width=\textwidth]{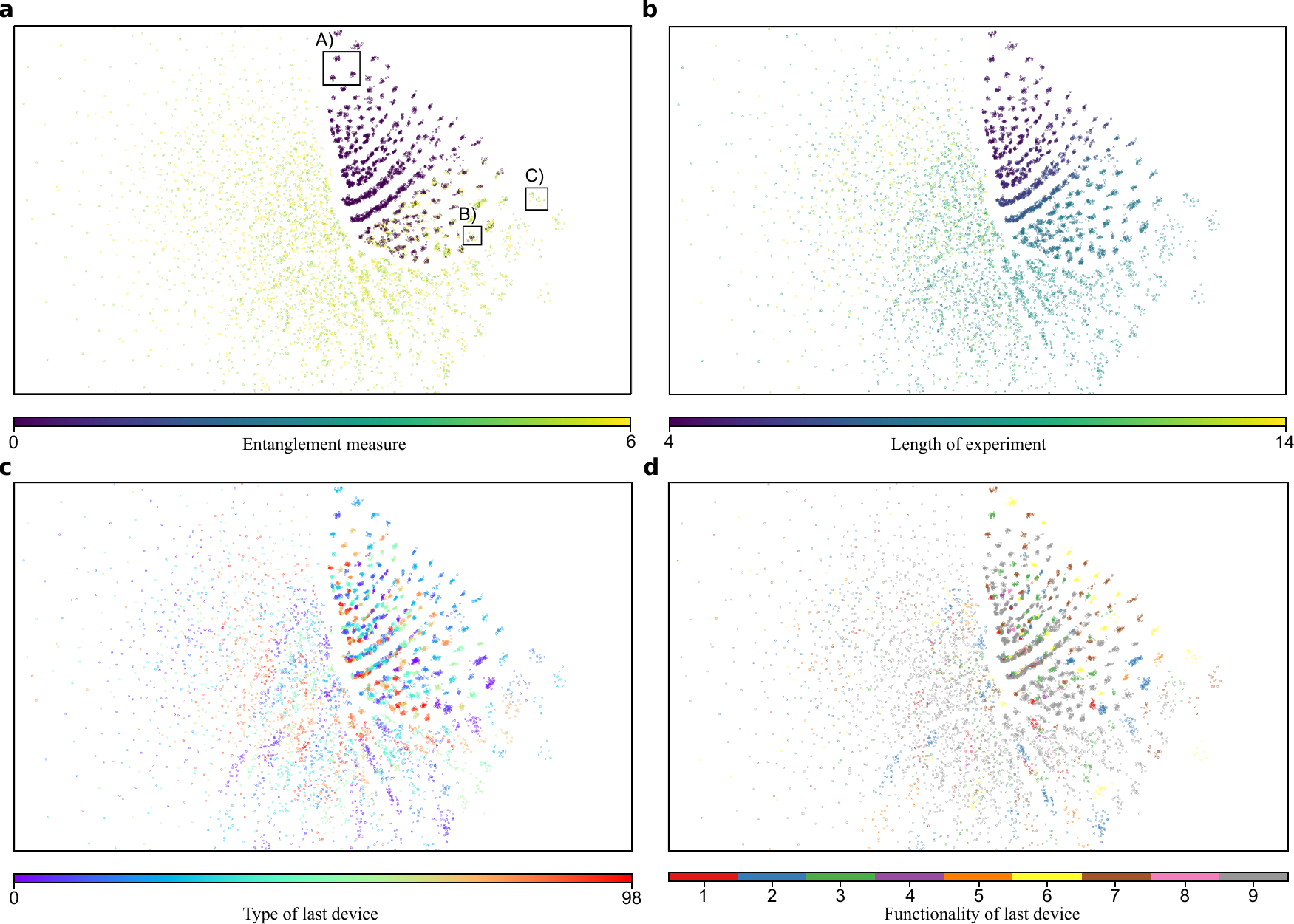}
\caption{\raggedright  \textbf{QOVAE Latent spaces.} \textbf{a-d} 2D latent space of the QOVAE-Low trained on equal numbers of $S=0$ and $0<S<7$ with $\ell<15$. Latent points colored \textbf{a)} by entanglement measure $S$, \textbf{b)} by length, \textbf{c)} by last devices, and \textbf{d)} Functionality of last device.}
\label{fig:latent1}
\end{figure*}

\begin{figure*}[t] \centering
\includegraphics[width=0.97\textwidth]{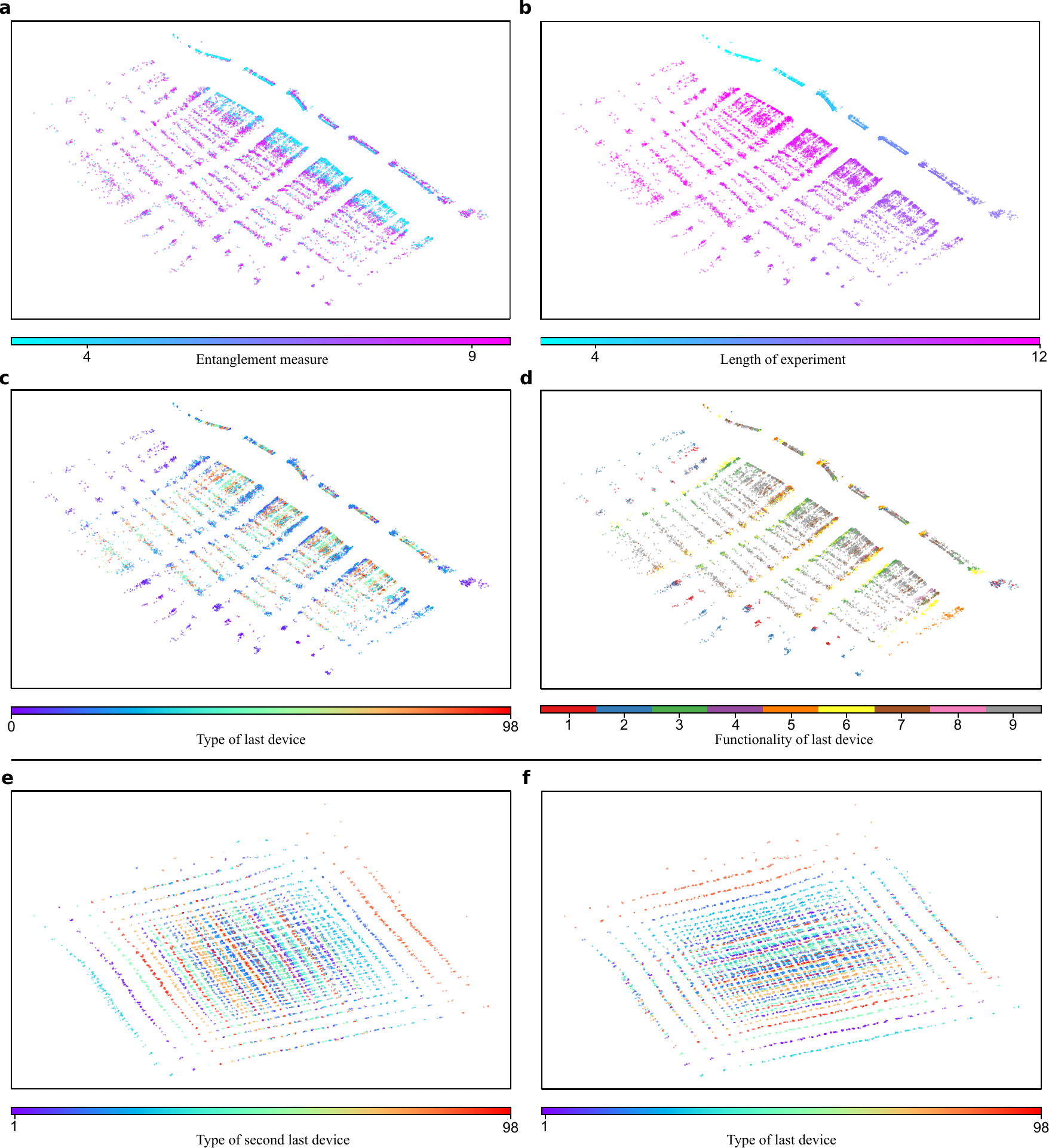}
\caption{\raggedright \textbf{More QOVAE Latent spaces.} \textbf{a-d} Latent spaces from a QOVAE trained on ~18K experiments with $3.0 < S < 4.0$ and $8.0 < S < 9.5$, and \textbf{e,f} Latent spaces from a QOVAE trained on 16K experiments with 10 devices. Latent points colored \textbf{a} by entanglement measure $S$, \textbf{b} by length, \textbf{c} by last devices, and \textbf{d} Functionality of Last device, as well \textbf{e} type of second last device \textbf{f} type of last device.}
\label{fig:latent2}
\end{figure*}

\textbf{6) The QOVAE learns an interpretable representation of quantum optics experiments.}\quad 

We train the QOVAE-Low on three different datasets of experiments and plot its learned latent space-- we discover that the QOVAE learns an intrepretible representation by structuring its learned representation according to the number and device order in its training experiments.

First we train the QOVAE-Low on equal distributions of $S=0$ and $0<S<7$ with $\ell<15$.
We directly plot the 2D latent space and interpret the latent space in terms of structure-property relations. 
We encode every training experiment into the latent space and color-code every latent point with its entanglement $S$. 
We observe interesting patterns and distinct clusters of setups including a region that is unentangled (colored in violet), and another that is entangled as well as some mixed regions (see Fig. \ref{fig:latent1}\textbf{a}. 
This is an interesting result, as the model is implicitly-- not directly doing this since it is not provided the entanglement label. 
Instead, it must be placing its latent space according to experiment structure. 

To understand exactly how, we analyse three different regions, one unentangled region A), one that shares entangled and unentangled states B) and one that contains only entangled states C).

In region A), we find three different clusters with no entangled states-- but all setups have only four devices, while in region B) and C), all setups have nine devices. 
We color-coding the latent space by setup length, as shown in Fig. \ref{fig:latent1}\textbf{b}, and find that the QOVAE encodes the training setups by number of optical devices. This shows a definitive correlation with entanglement. 

However, the size of the setup cannot alone be responsible for the latent structure learned. 
After all, region B) and C) have the same number of elements, but B) contains both entangled and unentangled setups, while C) contains only entangled ones. 

We analyse further and find, that the last element in every setup of region B) is always $\text{H}_b^{-2}$ while in C) it is always $\text{DC}_{ac}$. 
Therefore the QOVAE also structures its latent space by last setup element, we confirm this by color-coding all points by their final elements as shown in Fig. \ref{fig:latent1}\textbf{c}. 

This also explains why B) and C) have different entanglement $S$: in region C), that the last element is $\text{DC}_{ac}$ which is needed to create entanglement, 
while in region B), the last device contributes nothing to entanglement so these experiments are less likely to be entangled. 

We can further cluster all possible last elements into eight functional groups for instance, all holograms have the same effect on the entanglement property, so we can combine them to a single functional group. We plot the color-coded latent space according to functional groups in Fig. \ref{fig:latent1}\textbf{d} and see additional grouping. 

We conclude the QOVAE learns to structure its latent space according to experiment length and final device. 

It seems there was pronounced correlation between length and entanglement in the last training subset of experiments. 
We train another QOVAE-Low on entirely entangled states and show this is not the defining structure the QOVAE uses to learn. 
This time we use 18K experiments evenly split between low entanglement $3.0 < S < 4.0$ and high entanglement $8.0 < S < 9.5$. 
Similarly, in Fig. \ref{fig:latent2}, it is apparent that the QOVAE structured its latent space by length-- Fig. \ref{fig:latent2}\textbf{b} and 
last device type-- Fig. \ref{fig:latent2}\textbf{b} as before, however  there are clearer patterns and groups when plotting the functionality of last device type in Fig. \ref{fig:latent2}\textbf{d}. 

However, here we find that when we plot the entanglement in the latent space as shown in Fig. \ref{fig:latent2}\textbf{a} this does not correlate with the length of the experiments as plotted in Fig. \ref{fig:latent2}\textbf{b} or the last devices shown in Fig. \ref{fig:latent2}\textbf{c}. Interestingly, there are visible regions of low entanglement and high entanglement in the latent space that is thus unexplained by structure. Likely the QOVAE is learning some additional information about the sequence structure or order that is responsible for this. 

We further investigate how the QOVAE structures its latent space by removing length from the equation so the model can't use it-- in order to see additional insights into how the QOVAE structures its latent space and encodes the experiment sequence. We do this by training on 16K experiments that all have exactly 10 devices. 

We discover that the QOVAE doesn't structure its latent space using only the last device but will also use other information and devices in the experiment sequence, one clear example of this phenomenon is the latent space shown in Fig. \ref{fig:latent2}\textbf{e},\textbf{f}. 

Here, the QOVAE-Low uses not just the last device but also the second last device to structure its latent space and group them in rows that are perpendicular to each other. 
It is clear the QOVAE is using the dimensions of its latent space to store information about specific devices in the sequence ordering. 
The QOVAE-Low uses the two dimensions it has in order to encode the last and second last experiment devices in groups in the latent space. 
This means QOVAE's with higher dimensional latent spaces will be able to store more information about the devices at each location in the sequence ordering-- using  additional dimensions to encode and group devices. 

This description provides us with a complete interpretation of our model's latent representation and how it learns. 
With this new understanding, we can also explain results that we have observed in previous investigations. 
For example, we can explain exactly how the QOVAE can learn to generate from distributions of highly entangled states-- it simply learns the distribution of device orderings that define its training experiments. 

This distinguishes the QOVAE from other generative models applied to scientific domains-- by directly training on the device sequence, we enable 
the QOVAE to learn a human-interpretable representation that could be practically exploited to design experiments. 
In doing so we provide one of the very few examples of opening up and successfully demystifying black-box deep generative models applied to complex scientific systems.

\section*{Conclusions}
We presented the QOVAE which is the first deep generative model for the design of quantum optics hardware. Deep generative models are widely used but there has never been any investigation or understanding developed of their internal representation in a complex scientific domain. In a series of complex computational experiments, we investigated the QOVAE's \textit{internal picture of the quantum world}. The QOVAE was able to generate novel entangled experiments, learn distributions of entanglement and was shown to interpolate smoothly in its latent space-- which can also be used to search efficiently for new highly entangled experiments and target specific distributions of entangled states. When plotting the QOVAE's latent space we find complex internal structure, where the QOVAE learns a latent representation based on experiment length and device ordering. Our results go beyond designing new quantum optics-- they tackle the question of interpretability and explainability of black-box models  \cite{rudin2019stop, gilpin2018explaining} in a scientific domain. This is particularly promising since understanding what these model learn could lead to new computer-inspired scientific insights and discoveries.

\textbf{Optical devices.}\quad In order to ensure the device toolbox is suitable, all elements are used in standard quantum optics laboratories \cite{erhard2018experimental, erhard2020advances,pan2012multiphoton, krenn2017entanglement, malik2016multi}, including the technique of entanglement by path identity \cite{krenn2017entanglement}. 
In addition, we use a device toolbox that is similar to others that have been used in related work on machine learning assisted design of quantum optics experiments \cite{Krenn_2016, Melnikov_2018, adler2019quantum}. 

In our current approach, we use linear optics devices, however, it would be straightforward to include any other nonlinear optics devices like four-wave mixers \cite{wang2001generation} or optical parametric oscillators \cite{giordmaine1965tunable}.  In addition, it would be possible to train the QOVAE on a universal set of optical devices with continuous parameters by discretizing the space of those parameters. However, the performance of the QOVAE needs to be investigated when including such devices. Furthermore, the QOVAE could still produce experiments that are challenging to implement in the lab.

\textbf{Future work.}\quad In this paper, we  provide an initial demonstration of the QOVAE and its potential-- however, much further work needs to be done to truly assess and explore the use of its interpretable representation to investigate quantum optics experiments. Furthermore, while these investigations are exploratory, this work paves the way for many promising extensions that can be considered: conditional VAEs  \cite{kingma2014semi}, where the encoder and decoder would be trained on both the experiment and entanglement $S$ or other information about the quantum state in order to generate experiments conditionally. Another interesting extension is the hierarchical VAE, which has layers of latent variables \cite{sonderby2016ladder,zhao2017learning}-- these models have the ability to learn hierarchical features of the training data and have the potential to encode more information about experiment structure into their learned representation.
Using Bayesian optimization in the QOVAE's latent space it is possible to target many specific states but this is unlikely to be as efficient as domain specific methods \cite{krenn2020conceptual}. More interesting is how QOVAE learned to structure its latent  representation-- using its latent dimensions to independently encode the ordering of the experiment device sequence. Going forward, the most important future work is to figure out how this phenomenon can be exploited to develop new insights about quantum optics experiments and their design.  
Additionally, in future work, searching QOVAE's latent space, different metrics may be useful for different states-- for example like the GHZ state \cite{Bouwmeester_1999} or W states \cite{dur2000three} or any other specific experimental setups for high dimensionally entangled states \cite{erhard2020advances}. In place of fidelity $d_{\mathrm{FD}} = \left ( \sum_i\sqrt{p^*_i q_i } \right )^2$ mean squared error between states can be used:
$d_{\mathrm{SE}} = \sqrt{\sum_i (p^*_i - q_i )^2}$ where $p_i^*, q_i$ are basis ket probabilities of target and evaluated states. Another potential metric is the KL divergence between states
$d_{\mathrm{KL}} =  \sum_i p^*_i \log \frac{p^*_i}{q_i}$. 
It is also possible to target absolutely maximally entangled states \cite{cervera2019quantum, helwig2013absolutely} using the entanglement measure or the sum of the Schmidt rank vector components in place of fidelity.

\textbf{Other applications.}\quad Indeed the insight from this work are applicable to generative models of molecules \cite{kusner2017grammar, gomez2018automatic}. Molecules and quantum optics experiments are similar discrete, structured objects-- representable as graphs or sequences. 
If we treat quantum optics experiments are undirected we can even plot them as molecules by mapping devices to atoms-- see the supplementary for a few examples.  Based on how the QOVAE learns its representation we can infer that the ChemVAE \cite{gomez2018automatic} could learn a representation that is more human-interpretable by using its dimensions to store information about the ordering of the SMILES string \cite{weininger1988smiles}.
The QOVAE could, in principal, be directly be applied to other physical science domains, such as in the design of new quantum circuits for quantum computing. Currently, Noisy Intermediate-Scale Quantum (NISQ) computing algorithms \cite{preskill2018quantum,cerezo2021variational,bharti2021noisy} are promising candidates to surpasses the classical computational capabilities for numerous applications. Most of these approaches require good priors to explore efficiently and represent the space of parameters and solutions. The exponentially large Hilbert space formed by all possible quantum circuits makes this task computationally intractable when the structure-properties relation of these circuits is still not fully-understood. Only a few attempts to search this space have been made using genetic algorithms \cite{giraldi2004genetic, yabuki2000genetic, giraldi2004genetic, anand2021natural} which lack a sufficient prior. QOVAE's ability to learn meaningful representations as understood by domain experts could provide insights about how the Hilbert space is organized within these parameterized quantum circuits.
QOVAE learns an intrepretable representation of entanglement in quantum optics experiments. Our work with the QOVAE is an example in the physical sciences of opening the black-box of deep generative models to develop promising scientific insights.


\subsection*{Methods}

\subsubsection{Encoder} 

Our encoder is a diagonal Gaussian whose parameters are a mapping from the data manifold to the latent space. 
Our data, the quantum optics experiments $\mathbf x \in \mathbb{R}^{d \times T} $, are represented 
as a sequence with $T$ elements, each from a toolbox of $d$ possible devices,
\begin{align*}
q_{\phi} (\mathbf  z |\mathbf x) = \mathcal{N} ( \mathbf z \ | \ \bm \mu _{\phi}(\mathbf x),\bm \sigma^2 _{\bm \phi}(\mathbf x) ),     
\end{align*}
where $\bm \mu , \log \bm \sigma = g_{\bm \phi}(\mathbf x)$. 
First, $g$ consists of a convolutional neural network with 3 layers,
\begin{align*}
   \mathbf h = \text{Conv1d}_3(\text{Conv1d}_2(\text{Conv1d}_1(\mathbf x))).
\end{align*}
A single layer takes on the form
\begin{align*}
   \mathbf x' = \text{Conv1d}(\mathbf x) = \text{ReLU}(\mathbf w \odot \mathbf x+\mathbf b),
\end{align*}
where $\mathbf w \in \mathbb{R}^{n_f \times \ell \times d }$ is the convolution filter tensor which consists of $n_f$ filters each with length $\ell$ and $d$ features. Also the layer output is $\mathbf x' \in \mathbb{R}^{n_f\times T-\ell+1}$. 

For the input layer, this is just $d=D$ the number of devices in the toolbox used to create any experiment. 
ReLU(·) is the element-wise rectified linear unit function and
$\odot $ is the convolution operator which outputs a tensor with the following elements
\begin{align*}
(\mathbf w \odot \mathbf x)_{ft} = \sum _{l=1}^\ell \mathbf w_l ^f \cdot \mathbf x _{t+l-1},    
\end{align*}
where $\mathbf w_\ell ^f \in \mathbb{R}^{d}$ is the $f^{th}$ convolutional filter $\ell^{th}$ weight vector 
operating on the $t^{\text{th}}$ element (device) in the sequence (experiment). 

The second component of the encoder $g$ is a MLP with three layers that maps the flattened output $\mathbf h  $ from the convolutional neural net.  
to the parameters of latent distribution 
\begin{align*}
    \bm \mu , \ \ \log \bm \sigma =  \text{MLP}_g (\text{Flatten}(\mathbf h) ).
\end{align*}


\subsubsection{Decoder}

For the observation model, every data point is a sequence of devices from the $d$ element toolbox of possible device elements, 
thus we can model the data as independent categoricals whose mean vector is mapped from latent samples using a neural network,
\begin{align*}
    p_{ \bm \theta}(\mathbf  x  |\mathbf z) = \prod_{t=1}^T \text{Categorical} (\bm p_{\mathbf x_t}),
\end{align*}
where $\bm p_{\mathbf x_t} \in [0,1]^d$ is the probability vector of each device in the toolbox. Our encoder outputs these probabilities as
\begin{align*}
   \bm p_{ \mathbf x_1}, \dots , \bm p_{\mathbf x _t }, \dots ,\bm p_{\mathbf x_T }= f_{\bm \theta} (\mathbf z).
\end{align*}
First, $f$ consists of a three layer Recurrent neural network each with with Gated recurrent units (GRUs),
\begin{align*}
   \mathbf h = \text{GRU}_3(\text{GRU}_2(\text{GRU}_1(\text{MLP}_f(\mathbf z)))), 
\end{align*}
where $\text{MLP}_f$ is single layer MLP with a ReLU(.) activation. A single layer $\mathbf h  = \text{GRU}(x)$ takes on the form
\begin{align*}
  \mathbf z'_t &= \sigma (\mathbf W_i \mathbf x_t + \mathbf U_i \mathbf h_{t-1} +\mathbf b_i), \\
  \mathbf r_t &= \sigma (\mathbf W_r \mathbf x_t + \mathbf U_r \mathbf h_{t-1} +\mathbf b_r), \\
  \mathbf h'_t &= \text{Tanh} (\mathbf W_h \mathbf x_t + \mathbf U_h (\mathbf r_t \odot \mathbf h_{t-1}) +\mathbf b_h), \\
  \mathbf h_t &= (1-\mathbf z'_t) \odot \mathbf h_{t-1} + \mathbf  z_t' \odot \mathbf h'_t, 
\end{align*}
where $\mathbf W , \mathbf U, \mathbf b$ are parameters of the GRU layer. 
The $\mathbf z'_t$ and $\mathbf r_t$ are update and reset gate vectors, respectively, and
$\sigma $ is the sigmoid function.
The input to the first GRU layer is the $T$ length sequence of vectors $[\text{MLP}_f(\mathbf z), \dots , \text{MLP}_f(\mathbf z) ]$.
The output of the RNN is mapped to the device probabilities using a softmax layer
\begin{align*}
    \bm p_{\mathbf x_t} = \text{Softmax}(\mathbf W \mathbf h_t + \mathbf b) \ \ \ t=1 ,\dots T.
\end{align*}

\subsubsection{Training and data details} 

We manually performed an initial hyperparameter search for the QOVAE. 
We found an initial set of training hyperparameters using grid search.
We found training on around 1600 epochs to produce better validation accuracy and ELBO values. 
We trained the QOVAE using stochastic gradient descent and the Adam optimizer \cite{kingma2013auto} with a low learning rate $\sim 10^{-4}$.
Training was done using the KERAS machine learning package from tensorflow \cite{abadi2016tensorflow}. 
All models were trained on a V100 GPU node on the Beluga supercomputer. 
The minibatch size grid was $\{32, 64, 128, 256 \}$ with an optimal batch size of 64. 
The model architecture is the same for all experiments with both models.  
We use grid search to search over the encoder and decoder architecture. 
For the convolutional layers-- we search over $6,12,18,36$ filters with length $3,4,5$.
For the MLPs in both the decoder and encoder, we consider $32,64,128,256$ hidden units in each layer. 
For the GRU layer, we consider hidden states of size $128$, and a 64 unit layer that maps to the GRU layers. 

For new investigations, researchers can use the dataset provided to train the QOVAE, which has both $S>0$ and $S=0$ experiments. 
However, they can also produce new training datasets using the Melvin computer algorithm \cite{Krenn_2016} with a different device toolbox of interest. 
For this, the \texttt{sympy} symbolic algebra is required to specify how each device changes the state of the system \cite{codedata}. Melvin source code for creating new datasets is \href{https://github.com/XuemeiGu/MelvinPython}{available}.
After training the QOVAE they can investigate the learned representation by plotting the latent vectors for each training experiment in a 2 or 3 dimensional space. If the QOVAE has a higher dimensional latent space they can plot specific dimensions or project the space into 2 or 3 dimensions. Based on this, the researcher can investigate how the latent space represents the training entanglement or use it to search for specific states as they see fit-- either with random search or another search algorithm like the Bayesian optimization \cite{bo} approach we use.


\section*{Data Availability}
Training data is available: \\
\href{https://github.com/danielflamshep/qovae/blob/main/setups.smi}{https://github.com/danielflamshep/qovae/blob/main/setups.smi} \cite{codedata}.  

\section*{Code Availability}
Code is available: \\
\href{https://github.com/danielflamshep/qovae}{https://github.com/danielflamshep/qovae} \cite{codedata}. 

\section*{Acknowledgements}
A.A.-G. acknowledge support from the Canada 150 Research Chairs Program, the Canada Industrial Research Chair Program, and from Google, Inc. in the form of a Google Focused Award. M.K. acknowledges support from the FWF (Austrian Science Fund) via the Erwin Schr\"odinger fellowship No. J4309.

\section*{Author Contributions}

D.F.-S. conceived the overall project, developed the approach and wrote the paper. 
D.F.-S. and T.W. designed and performed investigations. 
A.C.L. provided technical advice. 
X.G. provided technical advice and wrote the entanglement calculation code.
M.K. built the dataset, provided technical advice and helped design the intrepretibility investigation and analysed experiments.
A.A.-G. led the project and provided overall directions. 
All authors participated in preparing the manuscript.

\section*{Competing Interests}
The authors declare that there are no competing interests.



\section*{references}

\section*{Supplementary Materials}

\begin{table}[t]
\resizebox{0.5\textwidth}{!}{
\begin{tabular}{c|c|c|c|c}
\hline 
\textbf{Device} & \textbf{Token} & \textbf{Visual} & \textbf{Operation} & \textbf{Operator}\\ 
\hline 
\\ 
\text{Down Conversion} & $ \texttt{DownConv} (\Psi, p, p')$ & &     
$ |\Psi\rangle + \sum_\ell |\ell\rangle_{p} |-\ell\rangle_{p'} $ & $\text{DC}_{pp'}$ \vspace{-0.8cm} \\ 
& & \includegraphics[width=0.1\columnwidth]{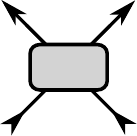} &&  \vspace{0.25cm} \\ 
\text{Beam Splitter} & $ \texttt{BS} (\Psi, p, p')$ &   &  
$ | \ell \rangle_{p}  \longrightarrow \frac{|\ell\rangle_{p'} +i|-\ell\rangle_{p}}{\sqrt{2}}  $ & $\text{BS}_{pp'}$ \vspace{-0.8cm} \\ 
& & \includegraphics[width=0.1\columnwidth]{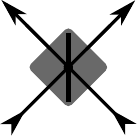} &&  \vspace{0.25cm}\\ 
\text{Mirror} & $ \texttt{Ref} (\Psi, p)$ &   &  
$ | \ell \rangle_{p}  \longrightarrow i|-\ell\rangle_{p} $ & $\text{R}_{p}$  \vspace{-0.8cm} \\ 
& & \includegraphics[width=0.1\columnwidth]{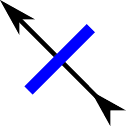} &&  \vspace{0.25cm}\\ 
\text{Dove Prism} & $ \texttt{DP} (\Psi, p)$ &   &  
$ | \ell \rangle_{p}  \longrightarrow ie^{i\pi \ell}|-\ell\rangle_{p} $& $\text{DP}_{p}$  \vspace{-0.8cm} \\ 
& & \includegraphics[width=0.1\columnwidth]{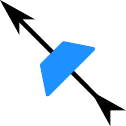} &&  \vspace{0.25cm}\\ 
\text{Hologram} & $ \texttt{OAMHolo} (\Psi, p, n)$ &   &  
$ | \ell \rangle_{p}  \longrightarrow |\ell+n\rangle_{p} $ & $\text{H}_{p}^{n}$ \vspace{-0.8cm} \\ 
& & \includegraphics[width=0.1\columnwidth]{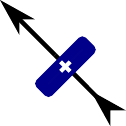} & &  \vspace{0.25cm}\\ 
\text{Hologram} & $ \texttt{OAMHolo} (\Psi, p, -n)$ &   &  
$ | \ell \rangle_{p}  \longrightarrow |\ell-n\rangle_{p} $ & $\text{H}_{p}^{-n}$ \vspace{-0.8cm} \\ 
& & \includegraphics[width=0.1\columnwidth]{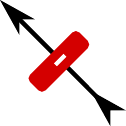} & &
\end{tabular}
} 
\caption{\raggedright The main table of devices divided into columns displaying 1) the standard device name 
    2) The token used in the sequence representation of the experimental setups and used to convert any setup to a sequence of one hot vectors
    3) The visual of the device used in the graph representation of the experimental setups
    4) the operator that the device represents that acts on the systems state
    5) the action the operator performs on any specific ket in the state}
\label{tab:dev}
\end{table}

\subsection{The quantum state}

\subsubsection{The quantum system}

We are investigating multipartite entanglement in a four photon system with two pairs of OAM-entangled photons. 
The state is represented by the OAM of the photon which is its orbital angular momentum. 
Each maximally entangled quantum optics experimental setup will produce some state 
which lives in the hilbert space defined by the tensor product of 
the individual subsystems defined by photons $a,b,c,d$ given by
$\mathcal{H} = \mathcal{H}_a \otimes \mathcal{H}_b \otimes \mathcal{H}_c \otimes \mathcal{H}_d $
\begin{align}
   \ket{\Psi} = \sum_{ijkl\in \ell_{\tiny\text{OAM}} }  \alpha_{ijkl} \ket{\psi _{ijkl} } 
\end{align}
where we can define a general state in the system as a superposition of the basis kets as defined 
\begin{align}
    \ket{\psi _{ijkl} } &= \ket{i}_a \otimes \ket{j}_b \otimes \ket{k}_c \otimes \ket{l}_d \\
                        &= \ket{i}_a\ket{j}_b \ket{k}_c\ket{l}_d = \ket{i,j,k,l}
\end{align}

where $i,j,k,l$ are the OAM quantum number of the photon, OAM states provide a suitable physical realisation 
of multilevel qudit systems which have been shown to improve the robustness of quantum key distribution schemes. 
In general, photon OAM states take on discrete integer values $m \in \mathbb{Z}$ with OAM $m\hbar$.
A proof-of-principle experiment with 7 OAM modes from $-3 \to   3$ has been demonstrated \cite{Mirhosseini_2015} .
\begin{align*}
    \ell_{\tiny\text{OAM}} = \{ -m_{-\ell}, -m_{-\ell} +1 , \dots , -1, 0, 1, \dots, m_\ell -1 , m_\ell  \}
\end{align*}

where $m_\ell$ is the maximal OAM number that can be reached by the setup, and $-m_{-\ell}$ is the smallest that can be reached.
This means each OAM quantum number can take on a possible $m_{-\ell}+m_\ell+1$ discrete values and since the system consists of four photons
($i\in \{a,b,c,d\}$)the dimension of the Hilbert space is $\prod_{i}( m_{-\ell_i}+m_{\ell_i}+1)$ and each basis ket lives in $\mathbb{C}^{\prod_{i} (m_{-\ell_i}+m_{\ell_i}+1)} $

\textbf{What states can be generated.} Based on the states produced in the training data using the devices in Table \ref{tab:dev} the OAM values don't go above $m_{-\ell_i} = m_{\ell_i}=12$ and the coefficients of basis kets can take on possible values :   
\begin{align*}
    \{k, kie^{i\pi m} \ | \ -m_{-\ell_i} \leq m \leq  m_{\ell_i} , \ m \in \mathbb{Z}\}
\end{align*}
where $k\in \mathbb{R}$ is a real prefactor from normalization.

\begin{figure}[t]
    \centering
    \includegraphics[width=0.5\columnwidth]{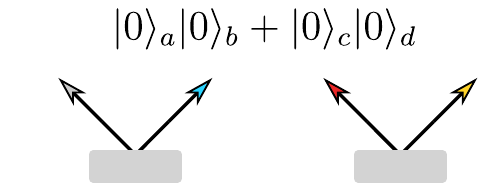}
    \caption{Two SPDC crystals create the initial state}
    \label{fig:my_label}
\end{figure}

\subsubsection{Initial State and SPDC}
Here, the initial state is created by a double spontaneous parametric down-conversion (SPDC) process. 
SPDC is a widespread source for the experimental generation of photon pairs. 
Multiple SPDC processes can be used to produce multipartite entanglement, 
as it is well known for the case of two-dimensional polarization entanglement \cite{Bouwmeester_1999, yao2012observation}.
However, instead of polarization, we are using the OAM of photons \cite{Allen_1992, Dada_2011, Romero_2012, Krenn_2014}, which is a discrete high-dimensional
degree of freedom based on the spatial structure of the photonic wave function

The input state of example 1 is a double-emission from SPDC, which leads to the 
initial state with general form :
\begin{eqnarray}
\ket{\psi}= N \left ( \sum_{\ell=-d_c}^{d_c} \ket{\ell}_a\ket{-\ell}_b \right )\otimes  \left ( \sum_{\ell=-d_c}^{d_c}  \ket{\ell}_c \ket{-\ell}_d \right )
\label{eq:SPDC1}
\end{eqnarray}
with $d_c$ being the highest order of SPDC considered, with photon pairs $a,b$ and $c,d$. $N$ is a normalization constant.  
During creation of the dataset we only consider $d_c=0$ with initial state 
\begin{align}
    \ket{\Psi} = \ket{0}_a\ket{0}_b \otimes  \ket{0}_c\ket{0}_d
\end{align}
This is the unnormalized state produced from the two initial SPDC devices, when we visualize the experiments we depict this as two grey rectangles to depict the crystal as in Figure 6.

\newpage 

\begin{algorithm}[t]
\caption{\texttt{State Calculation}}
\begin{algorithmic}[1]
\State \textbf{Input} $\B x  \sim p_{\text{data}}(\B x)$ 
\State \textbf{Initialize} \ $ \ket{\Psi_0} = \ket{0}_a\ket{0}_b + \ket{0}_c\ket{0}_d $
\State $ \mathcal{O}_1, \dots , \mathcal{O}_\ell \longleftarrow \B x $ 
\State $ \ket{\Psi} \longleftarrow  \mathcal{O}_\ell \cdots  \mathcal{O}_1 \ket{\Psi_0} $
\State $ \ket{\Psi} \longleftarrow \ket{\Psi} \otimes \ket{\Psi} $ 
\State $  \ket{\Psi} \longleftarrow \frac{\ket{\Psi}}{\braket{\Psi}{\Psi}} $
\State $\bm \rho  \longleftarrow\texttt{PartialTrace}(\ket \Psi) \texttt{ \footnotesize for all bipartitions}$ 
\State $\B s \longleftarrow -\text{Tr}(\bm \rho \log \bm \rho )$
\State $S = \sum_k s_k = \sum_{j} S(\rho_{aj})+ \sum_i S(\rho_i) $
\State \textbf{return} $S $ 
\end{algorithmic}
\end{algorithm}

\subsection{State and entanglement calculations}

To calculate the state of some experimental setup $\B x$ that has an initial state from the SPDC process
$ \ket{\Psi_0} = \ket{0}_a\ket{0}_b + \ket{0}_c\ket{0}_d $ 
we define the set of operations the state will undergo by extracting the operators defined from the sequence of optical devices that define the experiment. 
For each $\B x_t \in \B x$ we have a corresponding operator $ \mathcal{O}_t $ that changes the state by acting on it according to Table \ref{tab:dev}. 
Each device changes the state in the order defined by the sequence of the experiment and leads to a state $\ket{\Psi}$ which we then 
square collect four dimensionally entangled terms then normalize : $ \ket{\Psi} \otimes \ket{\Psi}/ \braket{\Psi}{\Psi}$. 
If there are no four particle or the state consists of one of the basis kets then the experiment is unentangled. 

Now we must quantify and calculate the entanglement in our system using the systems final state calculated in the previous paragraph and in Algorithm 1. 
Since our state lives in the hilbert space $\mathcal{H} = \mathcal{H}_A \otimes \mathcal{H}_B \otimes \mathcal{H}_C \otimes \mathcal{H}_D $ as 
\begin{align}
\ket{\Psi} = \sum_{ijkl} \alpha_{ijkl} \ket{i}_a\ket{j}_b \ket{k}_c\ket{l}_d
\end{align} 
the density matrix of the system is given by 
\begin{align}
\rho =  \ket{\Psi}\bra{\Psi} = \sum_{ijkl} \sum_{i'j'k'l'} \alpha_{ijkl}\alpha_{i'j'k'l'} \ket{ijkl}\bra{i'j'k'l'}
\end{align} 
We need to keep track of the four reduced density matrices for each subsystem or photon $\rho_a, \rho_b, \rho_c, \rho_d$. 
For example in the case of $a$ we can calculate $ \rho_a =\text{Tr}_{bcd}(\rho)  $ by tracing out the other subsystems. Explicitly :
\begin{align}
\rho_a = \sum_{jkl}  \bra{l}_d \bra{k}_c  \bra{j}_b  \cdot\ket{\Psi}  \bra{\Psi} \cdot \ket{j}_b  \ket{k}_c  \ket{l}_d 
\end{align} 
The other three can be calculated in similar fashion. We also need to keep track of the three reduced density matrices for each photon pair $\rho_{ab}, \rho_{ac}, \rho_{ad}$. 
Similarly in in the case of the photon pair $ab$ we can calculate $ \rho_{ab} =\text{Tr}_{cd}(\rho)  $ by tracing out the other subsystems. 
\begin{align}
\rho_{ab} = \sum_{kl}  \bra{k}_c \otimes \bra{l}_d \cdot\ket{\Psi}  \bra{\Psi} \cdot  \ket{k}_c \otimes \ket{l}_d 
\end{align} 
The other two can be calculated in similar fashion. 
we are interested in two main vectors quantifying the entanglement of the system ; 
1)  the von neumann entropy vector $\B s \in \R^{7}$ and 2) the Schmidt rank vector $\B r \in \Z^7$
where 
\begin{align}
\bm \rho  = \begin{bmatrix} \rho_a \\ \rho_b \\ \rho_c \\ \rho_d \\ \rho_{ab} \\ \rho_{ac} \\ \rho_{ad}\end{bmatrix} \ \ \
\B s = \begin{bmatrix} S(\rho_a) \\ S(\rho_b) \\ S(\rho_c) \\ S(\rho_d) \\ S(\rho_{ab}) \\ S(\rho_{ac}) \\ S(\rho_{ad})\end{bmatrix} \ \ \
\B r = \begin{bmatrix} \text{rank}( \rho_a) \\ \text{rank}(\rho_b) \\ \text{rank}(\rho_c) \\ \text{rank}(\rho_d) \\ \text{rank}(\rho_{ab}) \\ \text{rank}(\rho_{ac}) \\ \text{rank}(\rho_{ad})\end{bmatrix} 
\end{align} 

To explain the SRV, consider the simpler case of 3 particles and the state $\ket\Psi$ which has a SRV of (4,2,2) 
\begin{align}
    | \Psi \rangle = \frac{1}{2} (\ket{000} + \ket{101}+\ket{210}+\ket{311} )
\end{align}
Here, the first particle is four-dimensionally entangled with the other two parties,
whereas particle two and three are both only two dimensionally entangled with the rest. 
Also, $S(\cdot)$ is the von neumann entropy given by 
\begin{align}
    S(\rho_a) = -\text{Tr}(\rho_a \log \rho_a) = -\sum_s p_s \log p_s    
\end{align}

where $p_s$ are the eigenvalues of the quantum system $\rho_a$. 

Entanglement entropy is a measure of the degree of quantum entanglement between two subsystems constituting a two-part composite quantum system. 
Given a pure bipartite quantum state of the composite system, the reduced density matrix describes the state of a subsystem. 
The entropy of entanglement is the Von Neumann entropy of the reduced density matrix for any of the subsystems. 
If it is non-zero, the subsystem is in a mixed state and the two subsystems are entangled. 

We define our entanglement measure $S$ as the sum of all the entanglement entropies of all bipartitions of the system
\begin{align}
S = \sum_{j\neq a} S(\rho_{aj})+ \sum_i S(\rho_i)     
\end{align}
A bipartition of the system is a partition which divide the system into two parts $a$ and $b$, 
containing $n_1$ and $n_2$ particles respectively with $n_1+n_2=n$ supposing the quantum system consist of $n$ particles. 
Bipartite entanglement entropy is defined with respect to this bipartition.

\begin{figure*}[t]
    \centering
     \includegraphics[width=0.95\textwidth]{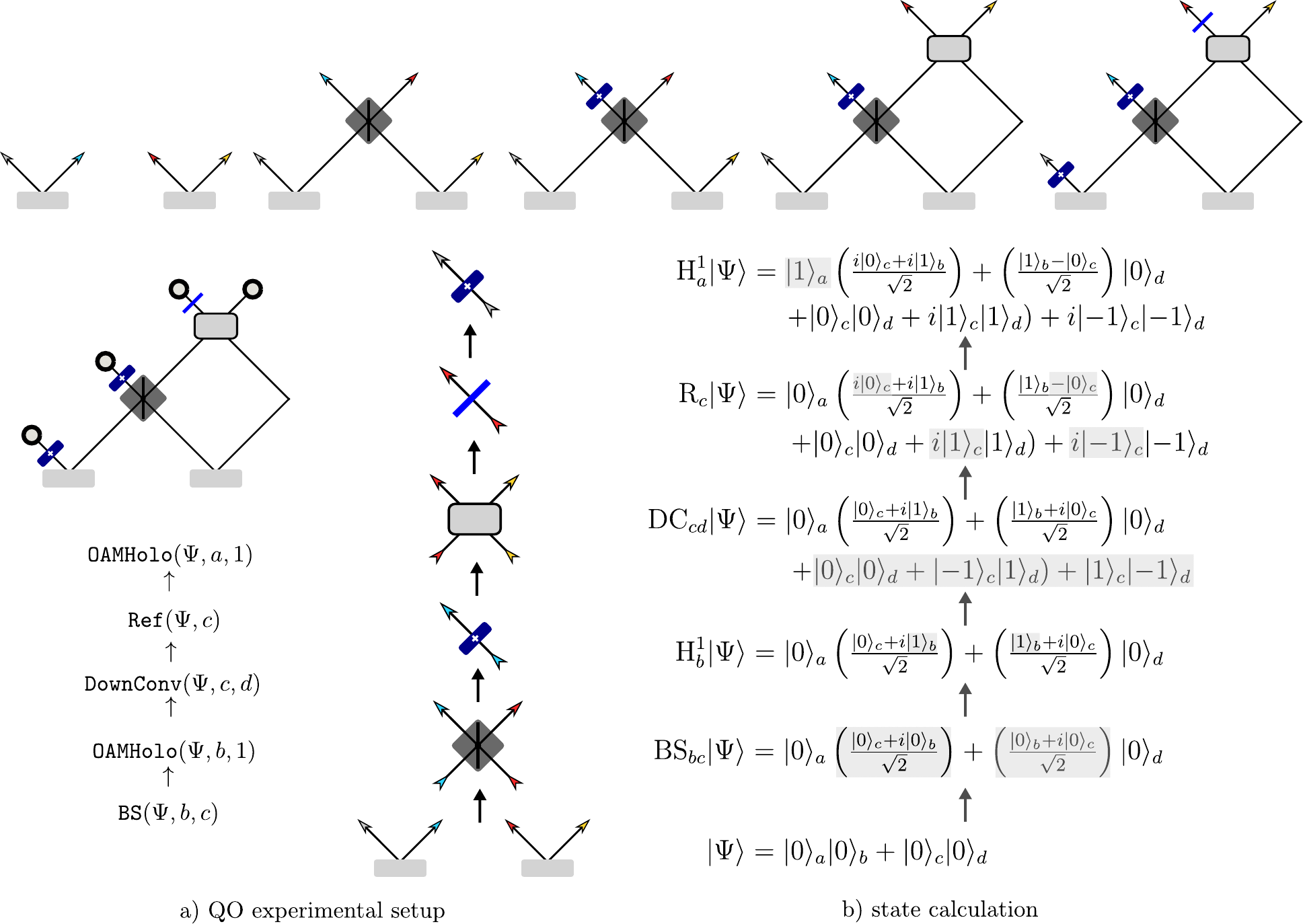}
     \caption{the graph of QO experimental setup whose state is calculated in the following section}
    \label{fig:exstate}
\end{figure*}

\subsection{Example state calculation}

The state calculation is automated using symbolic algebra from the sympy python package \cite{meurer2017sympy}. 
For demonstration purposes let's consider an example state calculation of a simple quantum optics experiment defined in 
Fig 8 the sequence of devices in the experiment operate on the state through the sequence of operators 
\begin{align}
 \text{BS}_{bc} \to    \text{H}_{b}^{1}\to    \text{DC}_{cd}  \to    \text{R}_{c}  \to    \text{H}_{a}^{1}     
\end{align}
starting with the initial state  $
    \ket{\Psi_0} = \ket{0}_a\ket{0}_b + \ket{0}_c\ket{0}_d
$
We apply each operator in order to find the final state given by 
\begin{align}
    \ket{\Psi} =   \text{H}_{a}^{1} \cdot \text{R}_{c} \cdot \text{DC}_{cd} \cdot \text{H}_{b}^{1} \cdot \text{BS}_{bc} \cdot \ket{\Psi_0}  
\end{align}
The first device in the setup is a beamsplitter on photon path b and c with operator $ \text{BS}_{bc} $ 
acting on the initial state replacing $b$ and $c$ kets with their superposition as   
\begin{align}
    \ket{0}_b \longrightarrow \bb{ \frac{\ket{0}_c+i\ket{0}_b}{\sqrt{2}} } \text{ and } 
    \ket{0}_c \longrightarrow \bb{\frac{\ket{0}_b+i\ket{0}_c}{\sqrt{2}}}
\end{align}
next the device $  \text{H}_{b}^{1} $ will add 1 OAM to all $b$ kets, so that the two zero OAM for photon b become $\ket{0}_b \longrightarrow \ket{1}_b $
Then applying device $\text{DC}_{c,d}$
\begin{align}
    \ket{\Psi} = \ket{\Psi}  + \ket{0}_c\ket{0}_d+\ket{-1}_c\ket{1}_d)+\ket{1}_c\ket{-1}_d
\end{align}
Then applying device $\text{R}_{c}$ and device $ \text{H}_a^1  $ we flip c and add a $i$ prefactor as well as increase the OAM of kets $a$. Then we square the state and normalize: 
$$\ket{\Psi} \to   \ket{\Psi} \otimes \ket{\Psi} \ \textit{and} \ \ket{\Psi} \to   \frac{\ket{\Psi}}{\braket{\Psi}{\Psi}}  $$
we are left with the following terms that contribute to 4 dimensional entanglement : 
\begin{align} \small
    \ket{\Psi} = \frac{1}{\sqrt{3}} (\ket{1,1,-1,-1} +  \ket{1,1,0,0} +   \ket{1,1,1,1} )
\end{align}

\subsection{Further experiments}

\begin{figure*}[h]
    \centering
    \includegraphics[width=\textwidth]{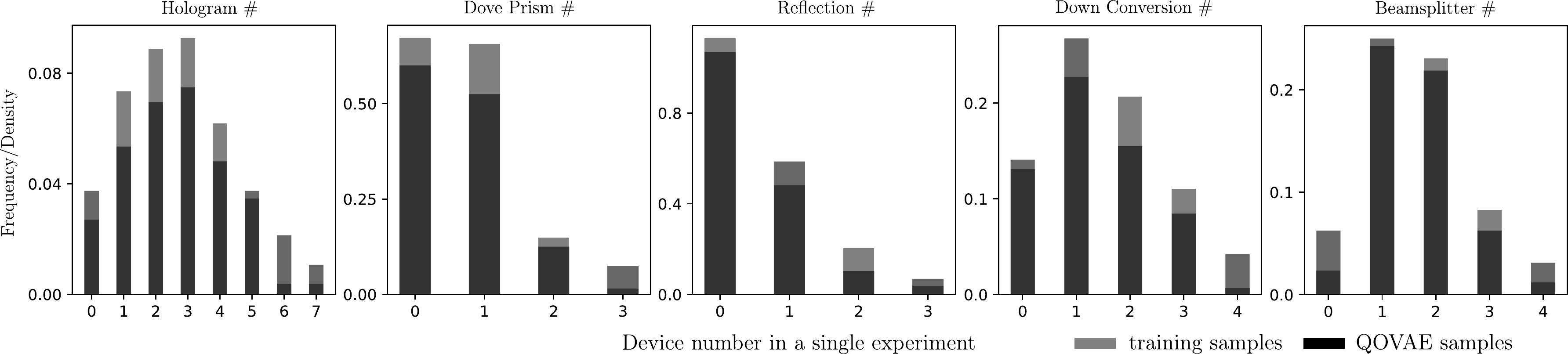}
    \caption{\raggedright \textbf{Learning structural distributions} 
    We check explicitly if the QOVAE-High has learned global properties of the training distribution. 
    We look at the distribution of devices in generated experiments to tell us what the model has learned about the explicit structure of training experiments.
By plotting histograms capturing the distribution of the number of basis element devices per experiment 
found from random samples of experiments from the training data and the QOVAE
we can see how well the QOVAE captured the structural distribution of the training data. 
To do so we sample 10k experiments from the QOVAE-High and its training data then plot histrograms of experiment device number. 
We focus on the main devices including number of holograms, dove prisms and reflection devices
as well as the number of double path devices like Down Converters and Beam-splitters. 
We can see that the model produces experiments with a similar average number of Dove Prisms and Reflection devices compared to the training experiments. Similarly, for the double path devices, the histograms are similar and the QOVAE leans them both reasonably well but underestimates slightly. For the hologram we also get a decent but not exact match. Overall, it is safe to say the QOVAE has learned the global structure in the training experiments.}
    \label{fig:dd}
\end{figure*}

\begin{figure*}[t]
    \centering
    \includegraphics[width=\columnwidth]{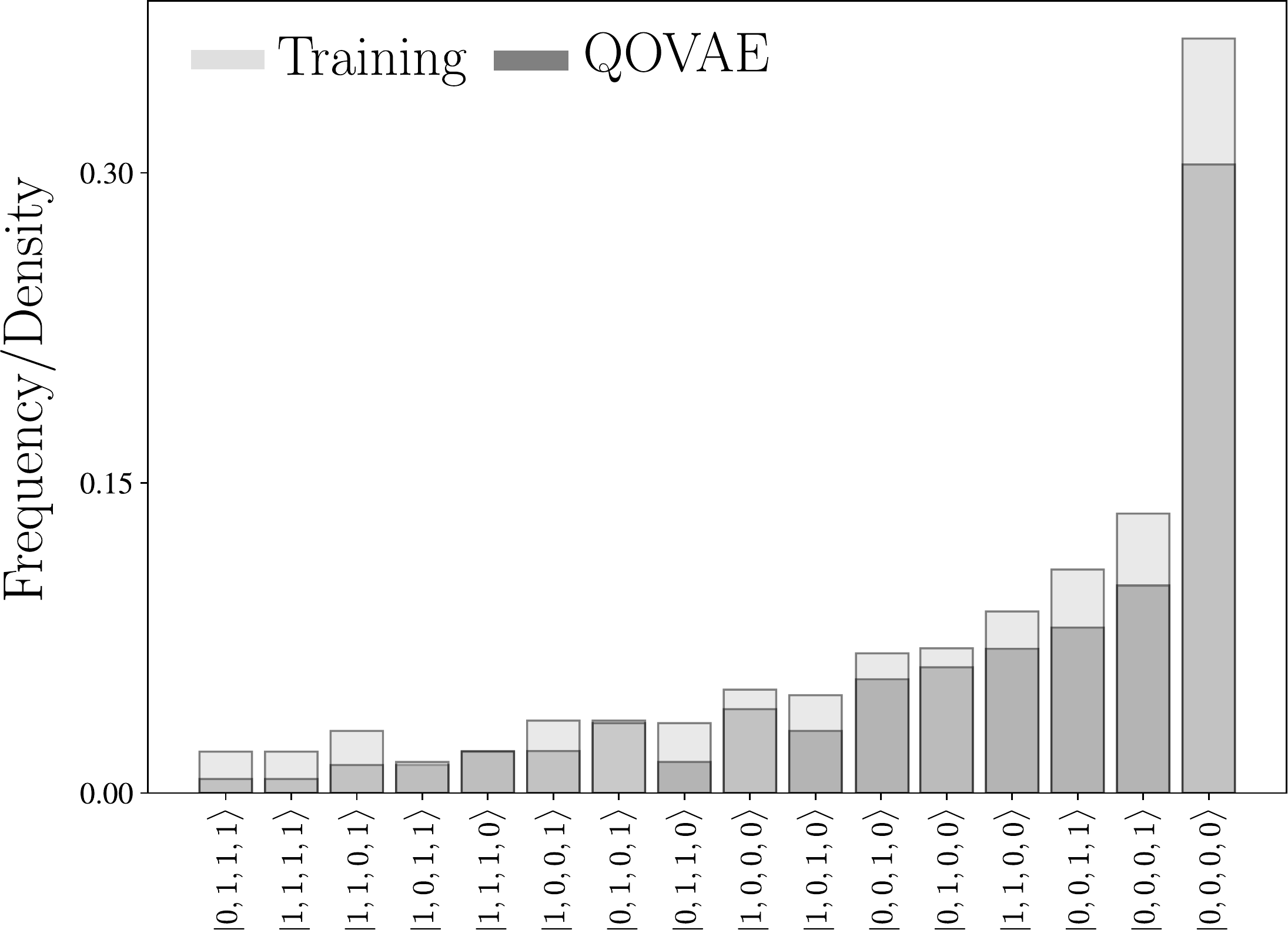}
    \caption{\raggedright \textbf{Distribution of States} To assess if the QOVAE-high learns a distribution of quantum states in the training data, we directly compare the frequency of basis kets $|i \rangle _a \otimes |j \rangle _b \otimes  |k \rangle _c \otimes |l \rangle _d $ arising in the states calculated from the 10K sampled experimental setups from the QOVAE and from the training data. To do this, here we plot histograms of basis kets with photon OAM mode either zero or one from both samples. It is clear that the basis kets with 0 or 1 OAM occur with the same frequency in quantum states produced by the QOVAE or states from the training experiments. }
    \label{fig:states}
\end{figure*}

\begin{table*}[t]
\begin{tabular}{l|c|c|c}
         & \multicolumn{3}{c}{Training data $S$ upper-bound} \\ 
       & $S<3$ & $S<4$ & $S<5$  \\  \hline \noalign{\smallskip}
        QOVAE $S$ 95 PT &  $ 5.21 \pm 0.32 $ & $ 5.66 \pm 0.36 $ &   $ 6.12 \pm 0.28 $  \\
        TRAIN \ $S$ 95 PT & $ 2.78 \pm 0.00 $ & $ 3.47 \pm 0.00 $ &   $ 4.45 \pm 0.00 $  \\ 
        \hline \noalign{\smallskip}
\end{tabular}
\caption{\raggedright \textbf{Learning larger entanglement than training.}
The 95th percentiles (PT) of $S$ from experiments from the QOVAE compared to training experiments for three different entangled datasets with upper-bounds of a) $S<3$ b) $S<4$ and c) $S<5$. Averages and standard errors are from 5 generations. The QOVAE's 95th percentile of $S$ is consistently larger than its training data. \red{This is likely the result of every experiment being represented as a one-hot encoded vector that is padded up to a maximum number of devices. This slightly biases the model to randomly sample larger experiments that are more likely to have higher $S$.} }
\label{tab:p}
\end{table*}



\begin{figure*}
    \centering
    \includegraphics[width=.99\textwidth]{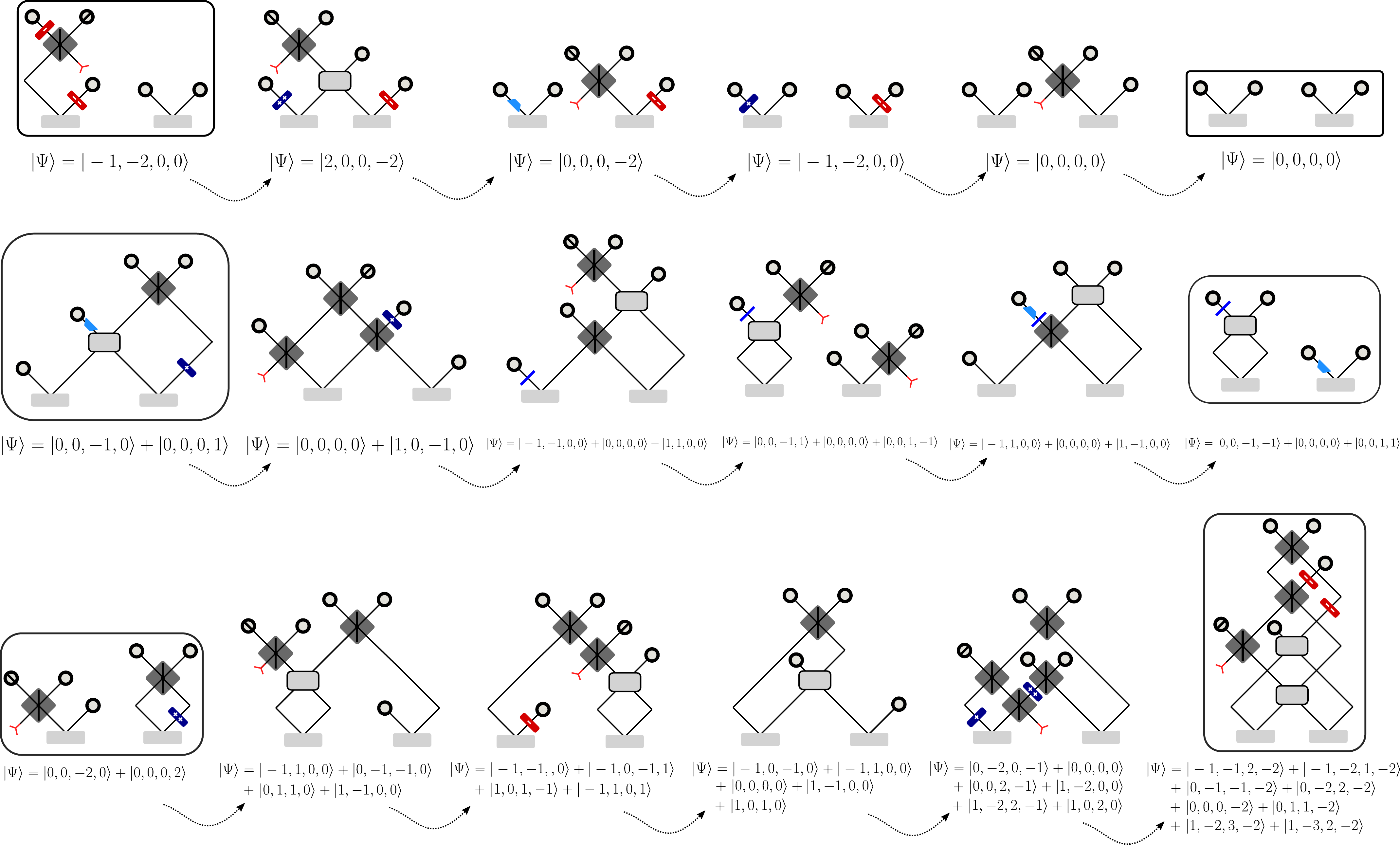}
    \caption{\raggedright \textbf{Interpolations between different quantum states} 
    To demonstrate the QOVAE learns a latent representation that encodes a measure of similarity of any experiment's
            quantum state we provide three interpolations between different states. The first using QOVAE-Low interpolates between two single basis ket's state. The next two from QOVAE-High : interpolate between a two ket state and three ket state as well as between a two ket state and eight ket state. For the first, we see that along the interpolation path the model decodes single ket states as well. For the second, we see that the interpolated kets have two then three kets. For the third, we see that the states gradually increase their ket number from two to eight.  }
    \label{fig:sint}
\end{figure*}

\begin{figure*}[t]
    \centering
\begin{tabular}{c|c}
 &  Samples  \\  \noalign{\smallskip} \hline \noalign{\smallskip}
\rotatebox{90}{\hspace{3cm} Training Samples}  & \includegraphics[width=0.9\textwidth]{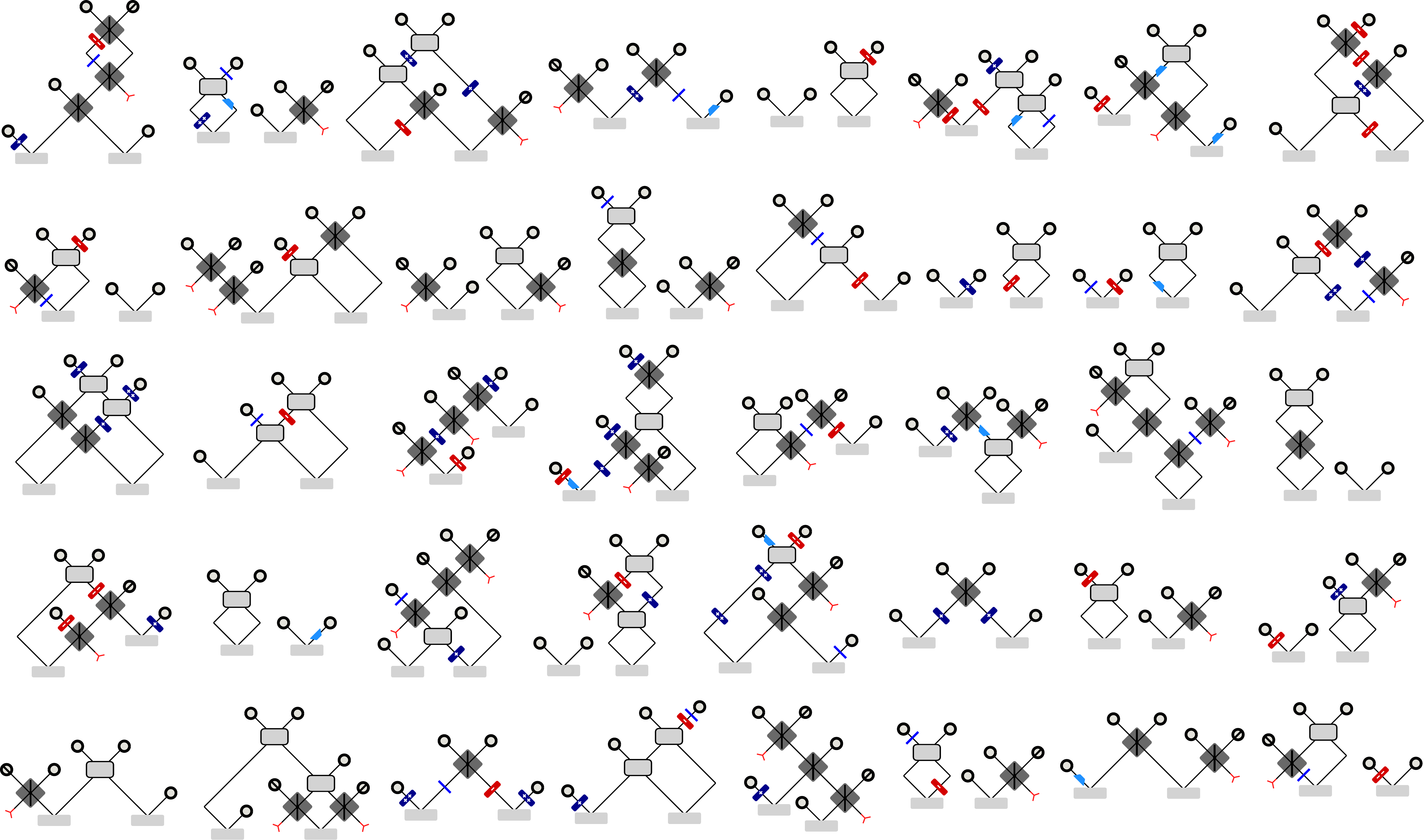}   \\ \noalign{\smallskip} \hline \noalign{\smallskip}
\rotatebox{90}{\hspace{3cm} QOVAE Samples}  &  \includegraphics[width=0.9\textwidth]{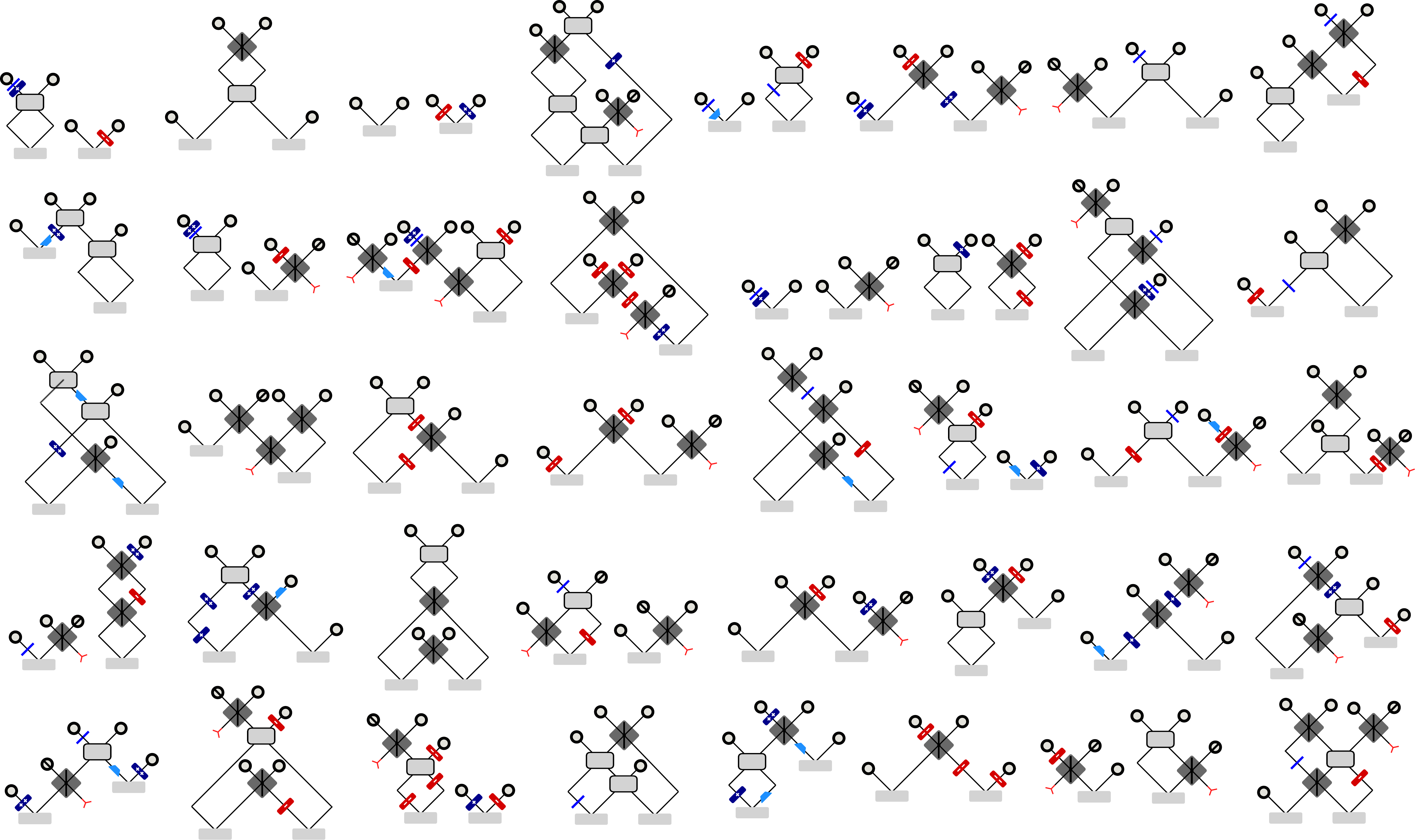}  \\  \noalign{\smallskip} \hline \noalign{\smallskip}  
\end{tabular}
\caption{\raggedright \textbf{More Samples.}  We random sample 40 experiments from the training data and from the QOVAE by sampling from the prior $\B z ^{(s)} \sim \mathcal{N} (\B 0, \B I)$ and passing that through the decoder $\{f(\B z ^{(s)})\}$. 
We display their graph representations. 
It is clear that both samples display a similar placement of devices and connectivity structures across the four possible photon paths.
}
\label{fig:my_label}
\end{figure*}

\end{document}